\documentclass[journal]{IEEEtran}
\usepackage{cite}
\ifCLASSINFOpdf
  \usepackage[pdftex]{graphicx}
\else
\fi
\usepackage{amsmath}
\usepackage{array}
\usepackage{bm}
\usepackage{amssymb}
\usepackage{color}
\usepackage{siunitx}
\usepackage[normalem]{ulem}
\usepackage{multirow}
\usepackage{comment}
\usepackage{cancel}

\newcommand{\nonfig}[1]{#1}

\DeclareRobustCommand{\Erase}{\bgroup\markoverwith{\textcolor{red}{\rule[.5ex]{2pt}{0.4pt}}}\ULon}
\begin{document}
%
% paper title
% Titles are generally capitalized except for words such as a, an, and, as,
% at, but, by, for, in, nor, of, on, or, the, to and up, which are usually
% not capitalized unless they are the first or last word of the title.
% Linebreaks \\ can be used within to get better formatting as desired.
% Do not put math or special symbols in the title.
\title{Smoothly Connected Preemptive Impact Reduction and Contact Impedance Control}
%
%
% author names and IEEE memberships
% note positions of commas and nonbreaking spaces ( ~ ) LaTeX will not break
% a structure at a ~ so this keeps an author's name from being broken across
% two lines.
% use \thanks{} to gain access to the first footnote area
% a separate \thanks must be used for each paragraph as LaTeX2e's \thanks
% was not built to handle multiple paragraphs
%

\author{Hikaru~Arita,~%\IEEEmembership{Member,~IEEE,}
  Hayato~Nakamura,
  Takuto~Fujiki,
  and~Kenji~Tahara%,~\IEEEmembership{Member,~IEEE}% <-this % stops a space
\thanks{
  H. Arita, H. Nakamura, T. Fujiki, and K. Tahara are with Department of Mechanical Engineering,
   Kyushu University, Japan,
  744 Motooka, Nishi-ku, Fukuoka, 819-0395, JAPAN (e-mail: arita@ieee.org; nakamura@hcr.mech.kyushu-u.ac.jp; fujiki@hcr.mech.kyushu-u.ac.jp; tahara@ieee.org).
  }% <-this % stops a space
% \thanks{Manuscript received April 19, 2005; revised August 26, 2015.}
}

\maketitle

% As a general rule, do not put math, special symbols or citations
% in the abstract or keywords.
\begin{abstract}
  This study proposes novel control methods that lower impact force by preemptive movement and smoothly transition to conventional contact impedance control.
  These suggested techniques are for force control-based robots and position/velocity control-based robots, respectively.
  Strong impact forces have a negative influence on multiple robotic tasks.
  Recently, preemptive impact reduction techniques that expand conventional contact impedance control by using proximity sensors have been examined.
  However, a seamless transition from impact reduction to contact impedance control has not yet been accomplished.
  It has been necessary to switch control laws or perform complicated parameter tuning.
  The proposed methods utilize a serial combined impedance control framework to solve the above problems.
  The preemptive impact reduction feature can be added to the already implemented impedance controller because the parameter design is divided into impact reduction and contact impedance control by using this framework.
  There is no undesirable contact force during the transition.
  Furthermore, even though the preemptive impact reduction employs a crude optical proximity sensor, the influence of reflectance is minimized using a virtual viscous force.
  Analyses and real-world experiments with the 1D mass model confirm these benefits.
  These benefits would be useful for many robots performing contact tasks.
\end{abstract}

% Note that keywords are not normally used for peerreview papers.

% For peer review papers, you can put extra information on the cover
% page as needed:
% \ifCLASSOPTIONpeerreview
% \begin{center} \bfseries EDICS Category: 3-BBND \end{center}
% \fi
%
% For peerreview papers, this IEEEtran command inserts a page break and
% creates the second title. It will be ignored for other modes.
\IEEEpeerreviewmaketitle

%%%%%%%%%%%%%%%%%%%%%%%%%%%%%%%%%%%%%%%%%%%%%%%%%%%%%%%%%%%%%%%%%%%%%%%%%%%%%%%%%%%
%% Contents 
%%%%%%%%%%%%%%%%%%%%%%%%%%%%%%%%%%%%%%%%%%%%%%%%%%%%%%%%%%%%%%%%%%%%%%%%%%%%%%%%%%%

%%%%%%%%%%%%%%%%%%%%%%%%%%%%%%%%%%%
\section{Introduction}
\label{sec:introduction}
%%%%%%%%%%%%%%%%%%%%%%%%%%%%%%%%%%%
% The very first letter is a 2 line initial drop letter followed
% by the rest of the first word in caps.
% 
% form to use if the first word consists of a single letter:
% \IEEEPARstart{A}{demo} file is ....
% 
% form to use if you need the single drop letter followed by
% normal text (unknown if ever used by the IEEE):
% \IEEEPARstart{A}{}demo file is ....
% 
% Some journals put the first two words in caps:
% \IEEEPARstart{T}{his demo} file is ....
% 
% Here we have the typical use of a "T" for an initial drop letter
% and "HIS" in caps to complete the first word.

% You must have at least 2 lines in the paragraph with the drop letter
% (should never be an issue)
%====================================%
\subsection{Background}
\label{sec:background}
%====================================%
\IEEEPARstart{T}{he} majority of robots require to manage transitions between contact and noncontact states.
Touching a robot hand to an object, landing or takeoff of legged or aerial robots, and collisions between a human or an environment and cooperative or mobile robots are examples of such transitions.
However, handling these transitions is well-known as one of the challenging problems in robotics.
The difficulties can be classified under two topics.
One is that the physical constraint changes dramatically during the transition; the robot's dynamic characteristic is discontinuous.
This phenomenon may make the robot go out of control.
The other is that an impact force is generated during the noncontact-to-contact transition.
Usually, an impact force is significantly large and thus dangerous.
The impact force, for example, may cause the robot and the environment to be damaged, cause the robot's hand to fail to grasp by a flick, knock the legged robot down, and unstabilize the robot control.

Typical conventional impact reduction methods use soft materials or force control.
Using soft materials is the standard approach for impact reduction\cite{alexander1990three,rond2020mitigating,giovanni2017design}.
Because of their flexibility, soft materials can absorb impact.
However, the absorbed massive force may cause the robot to oscillate.
Moreover, flexibility complicates accurate control, as mentioned in\cite{pajon2017walking}.
Despite active research in the field of soft robotics, these are known as difficult problems, i.e., the problem has only been replaced.

Force control has long been studied and is well known.
In particular, impedance control is practically used for manipulation, locomotion, and many other tasks\cite{park1999hybrid,ott2008on,samuel2009powered,lecours2012variable,ficuciello2015variable,huo2022impedance}.
Impedance control can reduce impact by setting parameters to make the desired impedance soft, and it can also perform dexterous tasks by making the desired impedance hard.
The impact reduction with force control, however, is limited because the response speed cannot be infinite.
The reduction method does not perform in time for the impulsive force having a higher frequency than the response frequency.

Preemptive impact reduction methods have recently been developed\cite{sato2022pre,guadarrama2022preemptive}.
These methods use proximity sensors mounted near the contact area and are based on impedance control.
Proximity sensors are short-range external sensors.
Optical proximity sensors have small sizes and fast responses compared to other types\cite{navarro2022proximity}.
Optical proximity sensors are attached to feet in these methods; moreover, the outputs are used to calculate virtual forces that serve as impedance control inputs.
The preemptive force input alleviates the problems mentioned above with classical impedance control and achieves high effects for impact reduction.
In these methods, however, an impedance control law works for both impact reduction and contact force control, i.e., these tasks interfere with each other.
Consequently, selecting parameters that consider overall performance\cite{sato2022pre} or switching between virtual and contact forces\cite{guadarrama2022preemptive} is required.
The former has a trade-off between impact reduction and contact force control performances, and the latter unstabilizes the system as with other switching methods.
The above transition between proximity and contact is listed in \cite{navarro2022proximity} as one of the problems attracting attention in the proximity perception field.

This study proposes novel control methods that perform preemptive impact reduction and contact force control.
A serial combined force control framework\cite{fujiki2021numerical} allows for a divided design for these tasks.
The transition from impact reduction to contact force control is smooth.
There is no switching of inputs and control laws.
The proposed techniques will improve the robotic performances of running, jumping, and other dynamic activities that have been limited to avoid damage caused by the impact force and will facilitate delicate works, such as grasping fragile objects and human interaction.
More specific characteristics of the proposed methods are highlighted in the following section compared to related works.

%=========================================%
\subsection{Related Works}
\label{sec:related_works}
%=========================================%
%-----------------------------------%
\subsubsection{(Contact) Impedance Control}
\label{sec:rel_contact_impedance}
%-----------------------------------%
\noindent
Impedance control is clusterized in force control-based type and position/velocity control-based type.
Force control-based impedance control\cite{hogan1985impedance} is called just ``impedance control,'' which calculates the force command based on the controlled object's position, velocity, and acceleration.
The controlled object is referred to as the ``plant'' in this study.
Position/velocity control-based impedance control\cite{kosuge1987mechanical} is called ``admittance control.'' 
The position, velocity, and acceleration of a virtual object are calculated from a force input in admittance control, and the virtual object is tracked by the position/velocity controller.
Although both controls aim to change the plant's dynamic behavior to the desired one, their characteristics differ.
Because the movement because of contact force is the input into impedance control, adequate back-drivability is required.
Furthermore, because impedance control uses a dynamic plant model to cancel itself through compensations such as a computed torque method, model errors, particularly friction, have an influence on performance, particularly positional accuracy.
Admittance control does not require high back-drivability because it collects information about contact force with a force sensor and the high gain position/velocity controller provides high positional accuracy.
However, no movement is generated by force applied to the area without a sensor.
Moreover, dramatically moving the virtual object because of a large contact force causes the system to become unstable.

Certain combined impedance and admittance controllers are proposed to incorporate both benefits.
Ott et al.\cite{ott2015hybrid} presented a parallel combined controller that, depending on the situation, switches between admittance and impedance control.
However, switching causes unstable behavior in general.
Fujiki and Tahara\cite{fujiki2021numerical} have developed a serial combined controller.
The admittance control part's calculated virtual object state is used as the impedance control part's equilibrium state.
The serial combined controller achieves higher stability than admittance control and higher positional accuracy than impedance control.
This study's proposed method employs the serial combined controller with generalization.

%-----------------------------------%
\subsubsection{Noncontact/Virtual/Preemptive Impedance Control}
\label{sec:rel_noncontact_impedance}
%-----------------------------------%
\noindent
Tsuji and Kaneko\cite{tsuji1999noncontact} proposed the first method for calculating virtual force from a noncontact sensor output and extending conventional impedance control to the noncontact region.
The method uses a vision sensor.
A similar method is used in \cite{lo2016virtual} to avoid collisions.

Because they are small, have fast responses, and can actively use reflection, optical proximity sensors have been used for impact reduction without occlusion.
Virtual force calculated from the proximity sensor's output was introduced in \cite{cheng2019comprehensive}.
The virtual force is an elastic force because its output depends on its distance from an object.

Sato et al.\cite{sato2022pre} proposed an impact reduction method using the virtual elastic force for legged robots.
The sensor is developed in such a manner that the virtual elastic force does not interfere with posture control after landing.
The control parameters are tuned by trial and error to maximize an index that includes impact reduction and posture control performances.
Because the control parameters influence both impact reduction and posture control performances, tuning is complex, and there is a trade-off.
In this study's proposed method, the parameters for impact reduction and contact impedance control are separated.
The separation simplifies the design of the controller according to a task.

Guadarrama-Olvera et al.\cite{guadarrama2022preemptive} proposed a method for using virtual elastic force as an input to a humanoid admittance controller.
By calculating the virtual wrench from the virtual elastic force, the method reduces impact and maximizes the landing contact area.
The virtual wrench is switched off after landing to prevent adding disturbance to the balance controller.
The parameter tuning considering the virtual wrench is not described.
They mentioned in the last section that achieving a smooth transition between virtual impedance control and conventional contact impedance control will help in improving walking stability.
The method proposed in this study achieves the above smooth transition from virtual impedance control to contact impedance control.
The results of analytical verification and experimental evaluation demonstrate the seamless property.

%-----------------------------------%
\subsubsection{Proximity Sensor-Based Control}
\label{sec:rel_proximity}
%-----------------------------------%
\noindent
Many methods for using proximity sensor outputs for position/velocity controllers have been proposed.
The method proposed by Koyama et al.\cite{koyama2019high} has particularly related to this paper.
The method uses virtual viscous force to slow the robot finger before contact.
The derivative of the distance output from the customized optical proximity sensor is used to calculate the virtual viscous force.
The virtual viscous force can be derived from the output ratio of a primitive optical proximity sensor.
Koyama et al.\cite{koyama2015grasping} demonstrated that the output ratio is not affected by reflectance.
In \cite{koyama2015grasping}, the output ratio estimates the time to contact for grasping with the robot finger.
Both methods in \cite{koyama2019high} and \cite{koyama2015grasping} stop the robot after contact.
As with other position/velocity-based control, switching is required if contact force control is to be done.
For impact reduction, the proposed method in this study employs the virtual viscous force derived from the output ratio of a primitive optical proximity sensor.

The method proposed by Arita and Suzuki\cite{arita2021contact} is one of the methods that use proximity sensors with force control.
To execute the desired transition between noncontact and contact, the method employs the virtual elastic force and changes its magnitude by controlling the proximity sensor's emitting light.
Although the transition is smooth, the desired contact force is constant, and the design of the desired force still requires to be investigated.
The contact force property of the method proposed in this study can be designed similarly to the conventional impedance control.

Ding and Thomas\cite{ding2021improving} proposed a method for adaptive parameter tuning of contact impedance control that makes use of proximity sensor outputs.
This method can prepare in advance for collisions that the collision avoidance function proposed in \cite{ding2020collision} cannot entirely avoid.
Because \cite{ding2021improving} aims to propose the preparation method, the transition between the collision avoidance feature and the preparation method is not mentioned, and impact forces are not evaluated.
In the future, the adaptive tuning method could be combined with the method proposed in this paper to improve the positional accuracy of the impedance control part.

%=========================================%
\subsection{Contents of This Paper}
\label{sec:contents}
%=========================================%
\noindent
This study proposes a controller for each force control-based and position/velocity control-based robot to achieve impact reduction and contact impedance control.
The impact reduction is preemptive and not easily affected by reflectance because it uses the virtual viscous force calculated from the optical proximity sensor output.
Using and generalizing the control framework proposed in \cite{fujiki2021numerical}, it is possible to combine impact reduction and contact impedance control.
This study first explains, analyzes, and evaluates the controller for force control-based robots because it is more naturally derived from \cite{fujiki2021numerical}.
The controller for position/velocity control-based robots is then presented, with the analytical results being the same for force control-based robots. 
Finally, a more generalized controller that includes the controllers proposed in \cite{fujiki2021numerical} and this paper is demonstrated.
To focus on collision, the plant is assumed to be a 1D mass model in this study.

The remainder of this study is structured as follows.
Section II introduces the proposed method for force control-based robots.
Section III investigates why the aforementioned method achieves theoretical independence of reflectance, divided design, and smooth transition.
Section IV evaluates the above method in actual experiments.
Section V contains the expansions, which include the proposed method for position/velocity control-based robots and the generalized controller.
Section VI describes the discussion of benefits, limitations, and future works.
Section VII concludes this paper.

% needed in second column of first page if using \IEEEpubid
%\IEEEpubidadjcol

%%%%%%%%%%%%%%%%%%%%%%%%%%%%%%%%%%%%%%%%%%%
\section{Proposed controller for force control-based robots}
\label{sec:proposal_force}
%%%%%%%%%%%%%%%%%%%%%%%%%%%%%%%%%%%%%%%%%%%
\nonfig{
  \begin{figure*}[!t]
    \centering
    \includegraphics[keepaspectratio,width=1.0\linewidth]{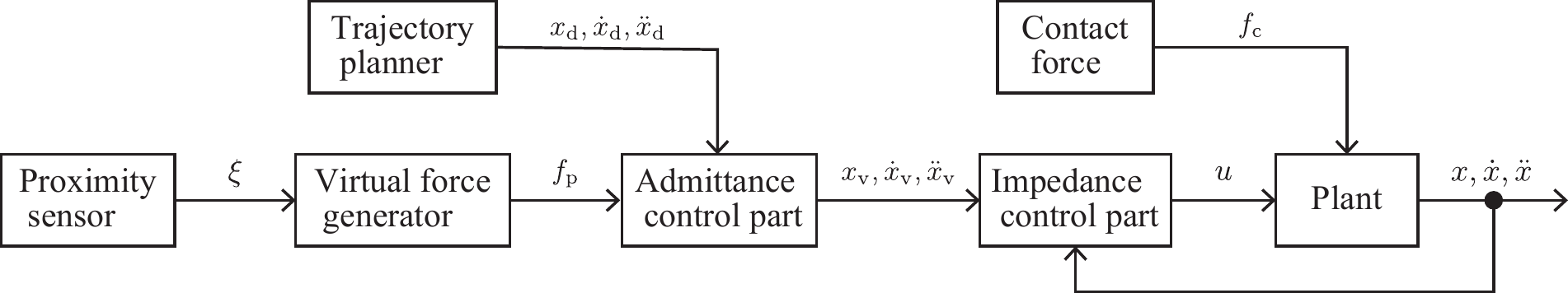}
    \caption{
      Block diagram of the proposed controller for force control-based robots.
      Serial combining admittance control and impedance control parts construct the controller.
      The unique point is that the force received by the admittance control part is not a contact force but a virtual force.
    }
    \label{fig:block_diagram_force}
  \end{figure*}
}
\noindent
Fig.~\ref{fig:block_diagram_force} shows the block diagram of the proposed controller for force control-based robots.
The primary differences between \cite{fujiki2021numerical} are that the forces in the impedance and admittance control parts are distinguished, and the force in the admittance control part is derived from the proximity sensor output.
This section briefly introduces the controller.
The detail is analyzed in the next section.

The trajectory planner determines the plant's desired state $(x_{\mathrm{d}}, \dot{x}_{\mathrm{d}}, \ddot{x}_{\mathrm{d}})$.
Because the planner should be designed as per a task, it is not the focus of this study.

A primitive optical reflective proximity sensor is used as the proximity sensor in Fig.~\ref{fig:block_diagram_force} in this study.
The minimum necessary components are a LED, a phototransistor, and two resistors.
The phototransistor generates the sensor output by receiving lights emitted by the LED and reflected at the surface of a detected object.

The virtual force generator in this study outputs the virtual viscous force~$f_{\mathrm{p}}$.
The equation is
\begin{equation}
    f_{\mathrm{p}} = G_{\mathrm{p}}\frac{\dot{\xi}}{\xi}, \label{eq:fp}
\end{equation}
where $\xi$ is the proximity sensor output, and $G_{\mathrm{p}}$ is the scale coefficient for the sensor output.
The subscript ``$\mathrm{p}$'' indicates that it is related to the proximity sensor.

The admittance control part simulates the virtual object affected by $f_{\mathrm{p}}$ with the following equation of motion.
\begin{equation}
    M_{\mathrm{a}} (\ddot{x}_{\mathrm{v}} - \ddot{x}_{\mathrm{d}}) + D_{\mathrm{a}}(\dot{x}_{\mathrm{v}} - \dot{x}_{\mathrm{d}}) + K_{\mathrm{a}}(x_{\mathrm{v}} - x_{\mathrm{d}}) = f_{\mathrm{p}}, \label{eq:dynamics_adm}
\end{equation}
where $M_{\mathrm{a}}$, $D_{\mathrm{a}}$, and $K_{\mathrm{a}}$ are the desired inertia, viscosity, and stiffness of the admittance control part, respectively, and $(x_{\mathrm{v}}, \dot{x}_{\mathrm{v}}, \ddot{x}_{\mathrm{v}})$ is the virtual object's state.
In conventional admittance control, the calculated state of the virtual object is tracked with a position/velocity controller.
For tracking, the control framework proposed in \cite{fujiki2021numerical} employs impedance control.

The impedance control part changes the dynamic behavior of the plant to the desired one.
The virtual object's state equals the equilibrium state of the desired dynamic behavior.
The plant's equation of motion is assumed as follows in this study.
\begin{equation}
    m\ddot{x} = u + f_{\mathrm{c}},\label{eq:dynamics_plant}
\end{equation}
where $m$ denotes the mass of the plant, $x$ denotes its position, $u$ denotes the controller's input force, and $f_{\mathrm{c}}$ denotes the contact force from an external environment or obstacle.
In this case, $u$ is given as follows:
\begin{align}
    u = &\left( \frac{m}{M_{\mathrm{i}}} - 1 \right) f_{\mathrm{c}} + m\ddot{x}_{\mathrm{v}} \nonumber\\
        &- \frac{m}{M_{\mathrm{i}}}\left[ D_{\mathrm{i}}(\dot{x} - \dot{x}_{\mathrm{v}}) + K_{\mathrm{i}}(x-x_{\mathrm{v}}) \right], \label{eq:input_imp}
\end{align}
where $M_{\mathrm{i}}$, $D_{\mathrm{i}}$, and $K_{\mathrm{i}}$ represent the desired inertia, viscosity, and stiffness of the impedance control part, respectively.
The plant's motion equation is modified by substituting \eqref{eq:input_imp} into \eqref{eq:dynamics_plant}:
\begin{equation}
    M_{\mathrm{i}} (\ddot{x} - \ddot{x}_{\mathrm{v}}) + D_{\mathrm{i}}(\dot{x} - \dot{x}_{\mathrm{v}}) + K_{\mathrm{i}}(x - x_{\mathrm{v}}) = f_{\mathrm{c}}. \label{eq:dynamics_imp}
\end{equation}
Even if \eqref{eq:dynamics_plant} is difficult, \eqref{eq:dynamics_imp} can be obtained by adding compensation terms to \eqref{eq:input_imp} as long as the model is provided.

$f_{\mathrm{p}}$ influences the plant's behavior via the virtual object's state.

%%%%%%%%%%%%%%%%%%%%%%%%%%%%%%%%%%%%%%%%%%%
\section{Analysis for force control-based}
\label{sec:analysis}
%%%%%%%%%%%%%%%%%%%%%%%%%%%%%%%%%%%%%%%%%%%
%=========================================%
\subsection{Independence of Reflectance}
\label{sec:independence}
%=========================================%
\noindent
This section confirms that \eqref{eq:fp} is, in theory, independent of reflectance.
A similar description is in \cite{koyama2015grasping}.

The output of an optical reflective proximity sensor $\xi$ can be modeled as follows:
\begin{equation}
    \xi = G_\xi\frac{\alpha\psi}{(d + d_{\rm{o}})^n},\label{eq:xi}
\end{equation}
where $G_\xi$ is the transform coefficient of the sensing element, such as a phototransistor, $\alpha$ is the detecting object's reflectance,
 $\psi$ is the energy of the light emitted from the LED, $d$ is the distance between the proximity sensor and the object,
  $d_{\rm{o}}$ is the offset distance to prevent direct contact between the electric components and the object, and $n$ is the diffusion coefficient fixed by the sensor design.
Only $d$ is time-varying.
$f_{\mathrm{p}}$ can be rewritten by substituting \eqref{eq:xi} into \eqref{eq:fp}:
\begin{equation}
    f_{\mathrm{p}} = -\frac{G_{\mathrm{p}}n}{d + d_{\mathrm{o}}}\dot{d}.\label{eq:fp2}
\end{equation}
As per \eqref{eq:fp2}, $f_{\mathrm{p}}$ is independent of $\alpha$, is proportional to $\dot{d}$, and is inversely proportional to $d$.
Therefore, $f_{\mathrm{p}}$ can be considered as a nonlinear virtual viscous force independent of reflectance.
When in contact, $f_{\mathrm{p}} = 0$ because $\dot{d}=0$ and the denominator is $d_{\mathrm{o}}$.

%=========================================%
\subsection{Divided Design}
\label{sec:divided_design}
%=========================================%
\noindent
This section explains that the parameters for impact reduction and contact impedance control are divided.
The Laplace transform is used in this section's analysis.

The Laplace transform of \eqref{eq:dynamics_adm} is as follows.
\begin{equation}
    (s^2M_{\mathrm{a}} + sD_{\mathrm{a}} + K_{\mathrm{a}})(X_{\mathrm{v}} - X_{\mathrm{d}}) = F_{\mathrm{p}},\label{eq:Laplace_A}
\end{equation}
where $X_{\mathrm{v}} = \mathcal{L}[x_{\mathrm{v}}]$, $X_{\mathrm{d}} = \mathcal{L}[x_{\mathrm{d}}]$, and $F_{\mathrm{p}} = \mathcal{L}[f_{\mathrm{p}}]$.
The following equation is obtained by solving \eqref{eq:Laplace_A} for $X_{\mathrm{v}}$.
\begin{equation}
    X_{\mathrm{v}} = X_{\mathrm{d}} + \frac{1}{s^2M_{\mathrm{a}} + sD_{\mathrm{a}} + K_{\mathrm{a}}}F_{\mathrm{p}}.\label{eq:Laplace_Xv}
\end{equation}
Similarly, solving \eqref{eq:dynamics_imp} for the $x$'s Laplace transform serves the following equation.
\begin{equation}
    X = X_{\mathrm{v}} + \frac{1}{s^2M_{\mathrm{i}} + sD_{\mathrm{i}} + K_{\mathrm{i}}}F_{\mathrm{c}},\label{eq:Laplace_X}
\end{equation}
where $X = \mathcal{L}[x]$, and $F_{\mathrm{c}} = \mathcal{L}[f_{\mathrm{c}}]$.
Eventually, $X$ is represented by substituting \eqref{eq:Laplace_Xv} into \eqref{eq:Laplace_X}:
\begin{equation}
    X = X_{\mathrm{d}} + \frac{1}{s^2M_{\mathrm{a}} + sD_{\mathrm{a}} + K_{\mathrm{a}}}F_{\mathrm{p}} + \frac{1}{s^2M_{\mathrm{i}} + sD_{\mathrm{i}} + K_{\mathrm{i}}}F_{\mathrm{c}}.\label{eq:Laplace_X_final}
\end{equation}
As per \eqref{eq:Laplace_X_final}, the influence on the plant's movement of $f_{\mathrm{p}}$ is determined only by parameters of the admittance control part, whereas the influence of $f_{\mathrm{c}}$ is determined only by parameters of the impedance control part.

%=========================================%
\subsection{Smooth Transition}
\label{sec:smooth_transition}
%=========================================%
\nonfig{
  \begin{figure}[!t]
    \centering
    \includegraphics[keepaspectratio,width=1.0\linewidth]{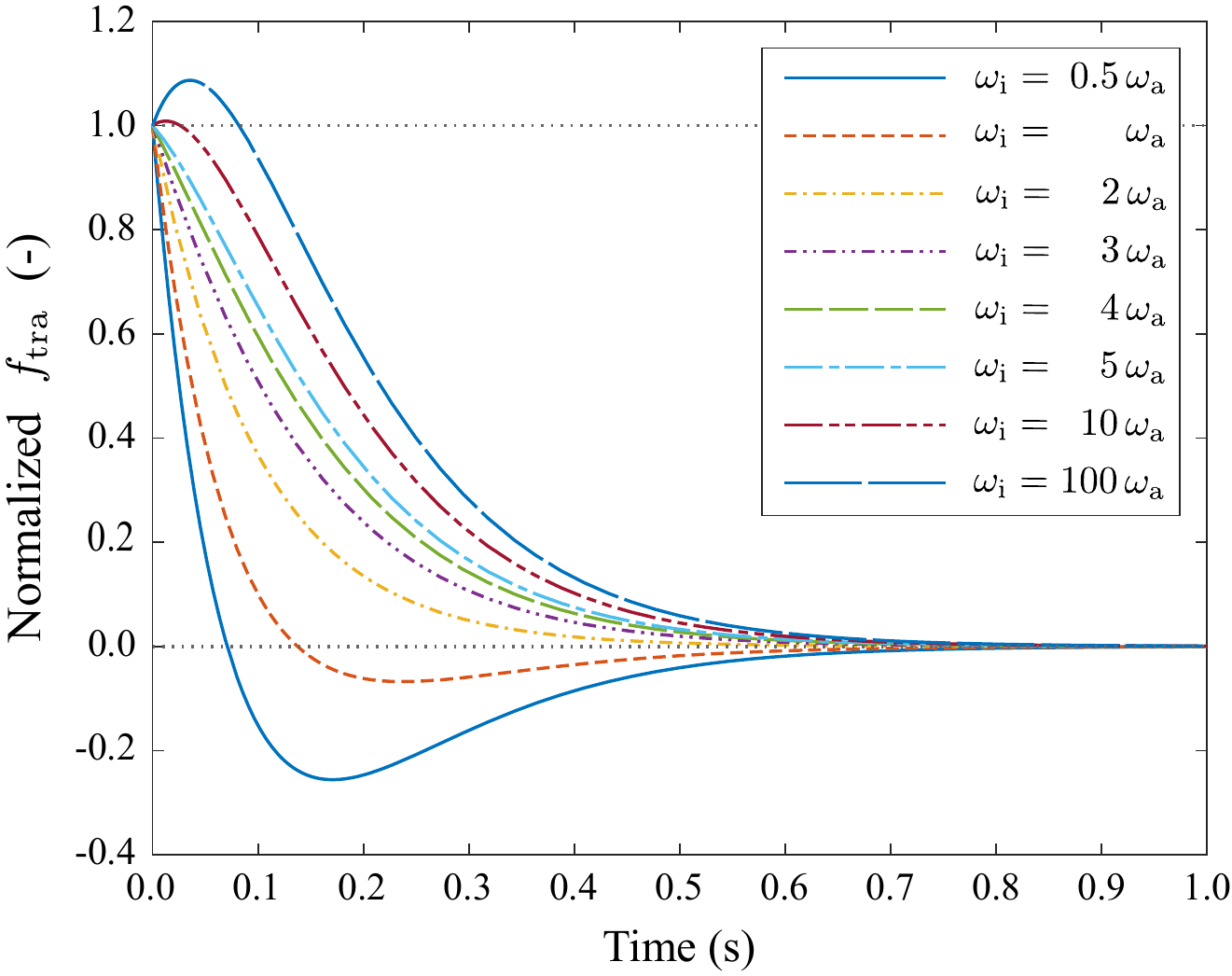}
    \caption{
      The relation between the time from the contact and $f_{\mathrm{tra}}(t)/f_{\mathrm{tra}}(0)$ $(\sigma=0.05, \nu=0.3, \omega_{\mathrm{a}} = 10, \zeta_{\mathrm{i}}=1).$
      Note that $f_{\mathrm{tra}}(0)<0$.
      There are parameter conditions that have the local maximum or minimum value before convergence.
    }
    \label{fig:graph_gen_ftra}
  \end{figure}
}

\nonfig{
  \begin{figure}[!t]
    \centering
    \includegraphics[keepaspectratio,width=1.0\linewidth]{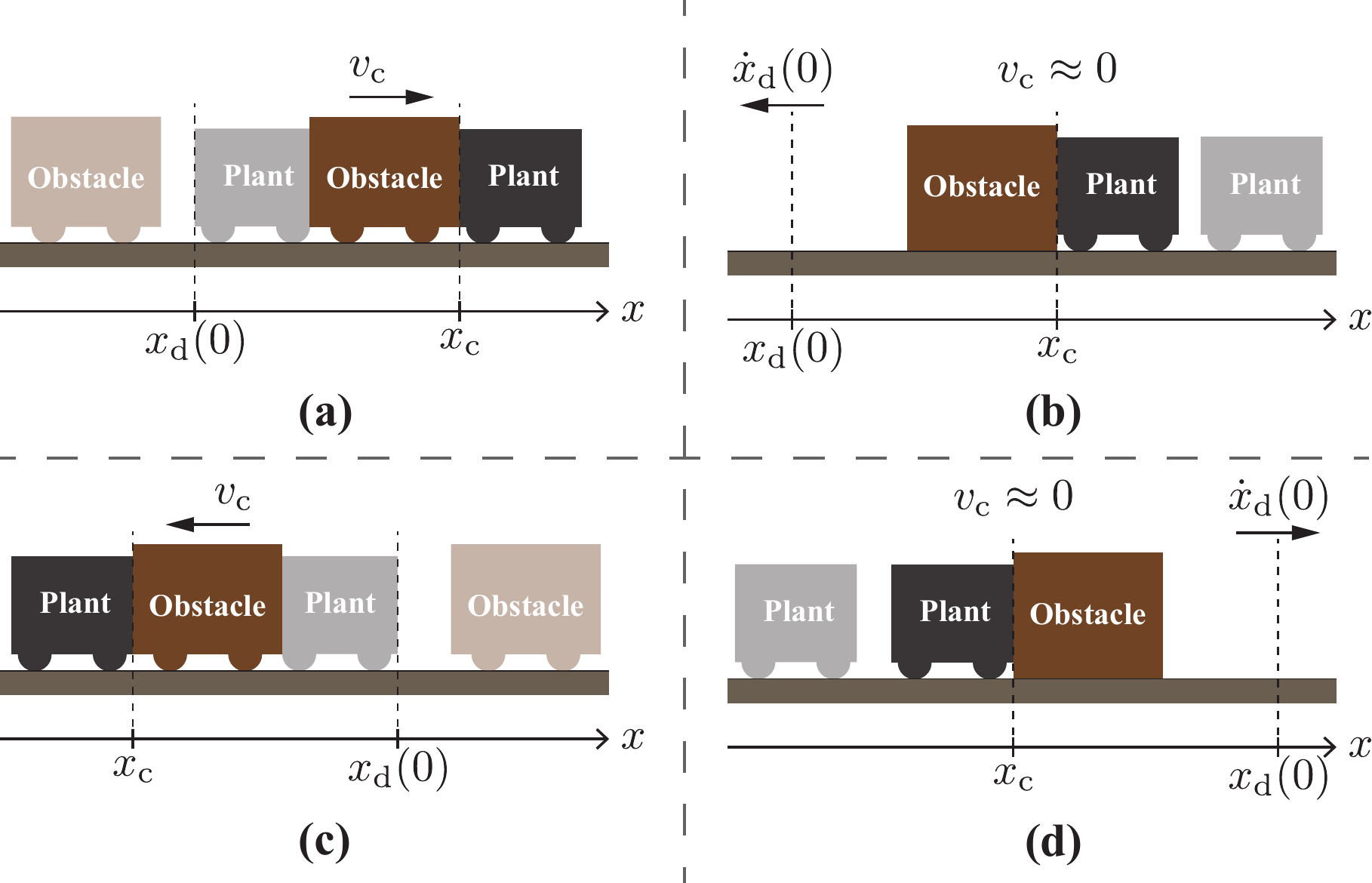}
    \caption{
      Situations at the moment of the contact. An obstacle approaches the plant in (a) and (c).
      The plant approaches an obstacle in (b) and (d).
      In (a) and (b), $\sigma>0, \nu>0, \eta>0$.
      In (c) and (d), $\sigma<0, \nu<0, \eta<0$.
    }
    \label{fig:situations}
  \end{figure}
}
\noindent
This section describes how selecting appropriate parameters results in the contact force immediately after the contact smoothly increases,
 and finally, contact impedance defined by the impedance control part is enabled.

Let $y = x_{\mathrm{v}} - x_{\mathrm{d}}$.
When in contact, \eqref{eq:dynamics_adm} can be expressed as follows because $f_{\mathrm{p}}=0$, as described in Section~\ref{sec:independence}.
\begin{equation}
    \ddot{y} + 2\zeta_{\mathrm{a}}\omega_{\mathrm{a}}\dot{y} + \omega_{\mathrm{a}}^2y = 0, \label{eq:eq_y}
\end{equation}
where $\zeta_{\mathrm{a}} =  D_{\mathrm{a}}/[2(M_{\mathrm{a}}K_{\mathrm{a}})^{1/2}]$, and $\omega_{\mathrm{a}} = (K_{\mathrm{a}}/M_{\mathrm{a}})^{1/2}$.
Let $t$ be the time from the contact.
If $\zeta_{\mathrm{a}}=1$, the general solution of \eqref{eq:eq_y} is
\begin{equation}
    y = (\eta t + \sigma)\exp(-\omega_a t), \label{eq:sol_y}
\end{equation}
where $\eta = \nu + \omega_{\mathrm{a}}\sigma$, $\sigma = y(0) = x_{\mathrm{v}}(0) - x_{\mathrm{d}}(0)$, and $\nu = \dot{y}(0)=\dot{x}_{\mathrm{v}}(0) - \dot{x}_{\mathrm{d}}(0)$.
Note that $\zeta_{\mathrm{a}}=1$ means \eqref{eq:dynamics_adm} is critical damping.
Equation \eqref{eq:dynamics_adm} decides the plant's movement after contact if $f_{\mathrm{c}}=0$ and transition property immediately after contact if $f_{\mathrm{c}}\neq 0$.
Critical damping is appropriate for \eqref{eq:dynamics_adm} because it has the fastest convergence without oscillation.
The transition property is explained further below.

The time derivatives of \eqref{eq:sol_y} are as follows.
\begin{align}
    \dot{y} &= (-\omega_{\mathrm{a}}\eta t + \nu)\exp(-\omega_{\mathrm{a}}t), \label{eq:sol_dy}\\
    \ddot{y} &= [\omega_{\mathrm{a}}^2 \eta t - \omega_{\mathrm{a}}(\eta+\nu)]\exp(-\omega_{\mathrm{a}}t). \label{eq:sol_ddy}
\end{align}
From \eqref{eq:dynamics_imp}, \eqref{eq:sol_y}, \eqref{eq:sol_dy}, and \eqref{eq:sol_ddy}, the following equation is derived.
\begin{align}
    f_{\mathrm{c}} = &M_{\mathrm{i}}(\ddot{x} - \ddot{x}_{\mathrm{d}}) + D_{\mathrm{i}}(\dot{x} - \dot{x}_{\mathrm{d}}) + K_{\mathrm{i}}(x-x_{\mathrm{d}}) \nonumber\\
                     &+\exp(-\omega_{\mathrm{a}}t)\{ -[M_{\mathrm{i}}\omega_{\mathrm{a}}^2 - D_{\mathrm{i}}\omega_{\mathrm{a}} + K_{\mathrm{i}}]\eta t \nonumber \\
                     &+ [M_{\mathrm{i}}\omega_{\mathrm{a}}(\eta + \nu) - D_{\mathrm{i}}\nu - K_{\mathrm{i}}\sigma]\}. \label{eq:sol_fc}
\end{align}
The initial contact force $f_{\mathrm{c}}(0)$ is determined by substituting $t=0$ into \eqref{eq:sol_fc}:
\begin{equation}
    f_{\mathrm{c}}(0) = M_{\mathrm{i}}(a_{\mathrm{c}} - \ddot{x}_{\mathrm{d}}(0)) + M_{\mathrm{i}}\omega_{\mathrm{a}}(\eta+\nu), \label{eq:init_fc}
\end{equation}
where $\ddot{x}(0) = \ddot{x}_{\mathrm{v}}(0) = a_{\mathrm{c}}$, $\dot{x}(0)=\dot{x}_{\mathrm{v}}(0) = v_{\mathrm{c}}$, and $x(0) = x_{\mathrm{v}}(0) = x_{\mathrm{c}}$. 
Note that the initial conditions assume that the tracking error caused by the impedance control part can be ignored until contact.
Equation \eqref{eq:Laplace_X} demonstrates that the assumption is met; however, the assumption is not strictly satisfied in the following analysis.
Although the assumption has an influence, the experimental results show that the amount is negligible in Section~\ref{sec:experiments}.
The first term of the right-hand side of \eqref{eq:init_fc} is the inertial force determined by the impedance control part.
The second term is affected by the initial position and velocity; even if the impact reduction is successful, the second term produces undesirable contact force.

The undesirable term can be eliminated by modifying the impedance control part.
$u$ is redefined by removing $m\ddot{x}_{\mathrm{v}}$ from \eqref{eq:input_imp}:
\begin{equation}
    u = \left( \frac{m}{M_{\mathrm{i}}} - 1 \right) f_{\mathrm{c}} - \frac{m}{M_{\mathrm{i}}}\left[ D_{\mathrm{i}}(\dot{x} - \dot{x}_{\mathrm{v}}) + K_{\mathrm{i}}(x-x_{\mathrm{v}}) \right]. \label{eq:input_imp2}
\end{equation}
The modification changes \eqref{eq:sol_fc} and \eqref{eq:init_fc} as follows:
\begin{align}
    f_{\mathrm{c}} = &M_{\mathrm{i}}\ddot{x} + D_{\mathrm{i}}(\dot{x} - \dot{x}_{\mathrm{d}}) + K_{\mathrm{i}}(x-x_{\mathrm{d}}) \nonumber\\
                     &+\exp(-\omega_{\mathrm{a}}t)[ (D_{\mathrm{i}}\omega_{\mathrm{a}} - K_{\mathrm{i}})\eta t - D_{\mathrm{i}}\nu - K_{\mathrm{i}}\sigma], \label{eq:sol_fc2}\\
    f_{\mathrm{c}}(0) = &M_{\mathrm{i}}a_{\mathrm{c}}. \label{eq:init_fc2}
\end{align}
$M_{\mathrm{i}}$ is frequently set to $m$ in practice to avoid the difficulty of measuring the contact force $f_{\mathrm{c}}$.
The above equations vary as follows in this case:
\begin{align}
    u = &-D_{\mathrm{i}}(\dot{x} - \dot{x}_{\mathrm{v}}) - K_{\mathrm{i}}(x-x_{\mathrm{v}}), \label{eq:input_imp3}\\
    f_{\mathrm{c}} = &m\ddot{x} + D_{\mathrm{i}}(\dot{x} - \dot{x}_{\mathrm{d}}) + K_{\mathrm{i}}(x-x_{\mathrm{d}}) \nonumber\\
                     &+\exp(-\omega_{\mathrm{a}}t)[ (D_{\mathrm{i}}\omega_{\mathrm{a}} - K_{\mathrm{i}})\eta t - D_{\mathrm{i}}\nu - K_{\mathrm{i}}\sigma], \label{eq:sol_fc3}\\
    f_{\mathrm{c}}(0) = &ma_{\mathrm{c}}. \label{eq:init_fc3}
\end{align}
The desired inertial force of the impedance control part is represented by the right-hand side of \eqref{eq:init_fc2} and \eqref{eq:init_fc3}.
In particular, \eqref{eq:init_fc3} shows the originally dynamic behavior of the plant without control.
Take note of the delay because the eliminated $m\ddot{x}_{\mathrm{v}}$ is the feedforward term of \eqref{eq:input_imp}.
Because the notation conveys the meaning, the inertia defined by the impedance control part is represented as $M_{\mathrm{i}}$ in the following analysis.

Equation \eqref{eq:sol_fc2} is the sum of the desired equation of motion of conventional impedance control and the fourth term on the right-hand side.
The fourth term is extracted from \eqref{eq:sol_fc2} as follows.
\begin{equation}
    f_{\mathrm{tra}} := \exp(-\omega_{\mathrm{a}}t)[ (D_{\mathrm{i}}\omega_{\mathrm{a}} - K_{\mathrm{i}})\eta t - D_{\mathrm{i}}\nu - K_{\mathrm{i}}\sigma]. \label{eq:ftra}
\end{equation}
The initial and final values are
\begin{align}
    f_{\mathrm{tra}}(0) &= - D_{\mathrm{i}}\nu - K_{\mathrm{i}}\sigma,\\
    f_{\mathrm{tra}}(\infty) &= 0.
\end{align}
It is desirable that $f_{\mathrm{tra}}(0)<f_{\mathrm{tra}}(t)<f_{\mathrm{tra}}(\infty)$ or $f_{\mathrm{tra}}(\infty)<f_{\mathrm{tra}}(t)<f_{\mathrm{tra}}(0)$ is satisfied.
However, as shown in Fig.~\ref{fig:graph_gen_ftra}, $f_{\mathrm{tra}}$ may exceed the range.
The following part analyzes the extreme value of $f_{\mathrm{tra}}$ to identify the condition satisfying the above range.

Because the extreme value analysis includes the evaluation of inequalities, the initial state signs must be confirmed.
The situations to be possible to occur are shown in Fig.~\ref{fig:situations}.
Fig.~\ref{fig:situations} shows the situation when an obstacle approaches the plant and the one when the plant approaches an obstacle.
In each situation, $\sigma$, $\nu$, and $\eta$ have the same sign, and the analytical results of all situations are equivalency.
The signs of $\sigma$, $\nu$, and $\eta$ are positive in the following analysis.

The time derivatives of \eqref{eq:ftra} are as follows.
\begin{align}
    \dot{f}_{\mathrm{tra}} = &\exp(-\omega_{\mathrm{a}}t)[ -\omega_{\mathrm{a}}(D_{\mathrm{i}}\omega_{\mathrm{a}} - K_{\mathrm{i}})\eta t \nonumber\\
                             &+(D_{\mathrm{i}}\omega_{\mathrm{a} - K_{\mathrm{i}}})\eta +\omega_{\mathrm{a}}(D_{\mathrm{i}}\nu + K_{\mathrm{i}}\sigma)], \label{eq:dftra}\\
    \ddot{f}_{\mathrm{tra}} = &\exp(-\omega_{\mathrm{a}}t)[ \omega_{\mathrm{a}}^2(D_{\mathrm{i}}\omega_{\mathrm{a}} - K_{\mathrm{i}})\eta t \nonumber\\
                             &-2\omega_{\mathrm{a}}(D_{\mathrm{i}}\omega_{\mathrm{a} - K_{\mathrm{i}}})\eta -\omega_{\mathrm{a}}^2(D_{\mathrm{i}}\nu + K_{\mathrm{i}}\sigma)]. \label{eq:ddftra}
\end{align}
Let $t_{\mathrm{ex}}$ be the time when $\dot{f}_{\mathrm{tra}}=0$.
$t_{\mathrm{ex}}=\infty$ or 
\begin{equation}
    t_{\mathrm{ex}} = \frac{1}{\omega_{\mathrm{a}}} + \frac{D_{\mathrm{i}}\nu + K_{\mathrm{i}}\sigma}{(D_{\mathrm{i}}\omega_{\mathrm{a}}- K_{\mathrm{i}})\eta}. \label{eq:tex}
\end{equation}
The case when $t_{\mathrm{ex}}=\infty$ is trivial.
The case of \eqref{eq:tex} is analyzed below.

From \eqref{eq:ddftra} and \eqref{eq:tex}, the following equation is obtained.
\begin{align}
    \ddot{f}_{\mathrm{tra}}(t_{\mathrm{ex}}) = &-(D_{\mathrm{i}}\omega_{\mathrm{a}} - K_{\mathrm{i}}) \nonumber\\
                                                &\times \omega_{\mathrm{a}} \eta \exp\left( -1-\frac{D_{\mathrm{i}}\nu + K_{\mathrm{i}}\sigma}{\omega_{\mathrm{a}}(D_{\mathrm{i}}\omega_{\mathrm{a}}-K_{\mathrm{i}})\eta} \right).
\end{align}
Because $\omega_{\mathrm{a}}>0$, $\eta>0$, and $\exp(\cdot)>0$, 
\begin{equation}
    \mathrm{sgn}(\ddot{f}_{\mathrm{tra}}(t_{\mathrm{ex}})) = -\mathrm{sgn}(D_{\mathrm{i}}\omega_{\mathrm{a}} - K_{\mathrm{i}}).\label{eq:ddtra_tex}
\end{equation}

From \eqref{eq:ddtra_tex}, $f_{\mathrm{tra}}$ has a local maximum value when $D_{\mathrm{i}}\omega_{\mathrm{a}} - K_{\mathrm{i}}>0$, i.e.,
\begin{equation}
    \omega_{\mathrm{i}} < 2\zeta_{\mathrm{i}}\omega_{\mathrm{a}}, \label{eq:omega_i_condition1}
\end{equation}
where $\omega_i = (K_{\mathrm{i}}/M_{\mathrm{i}})^{1/2}$, and $\zeta_{\mathrm{i}} = D_{\mathrm{i}}/[2(M_{\mathrm{i}}K_{\mathrm{i}})^{1/2}]$.
In this case, because the second term on the right-hand side of \eqref{eq:tex} is positive, $t_{\mathrm{ex}}$ is always positive.
Consequently, if \eqref{eq:omega_i_condition1} is met, $f_{\mathrm{tra}}$ has a local maximum value after contact.

On the other hand, $f_{\mathrm{tra}}$ has a local minimum value when $D_{\mathrm{i}}\omega_{\mathrm{a}} - K_{\mathrm{i}}<0$, i.e.,
\begin{equation}
    \omega_{\mathrm{i}} > 2\zeta_{\mathrm{i}}\omega_{\mathrm{a}}. \label{eq:omega_i_condition2}
\end{equation}
Because the second term on the right-hand side of \eqref{eq:tex} is negative in this case, $t_{\mathrm{ex}}$ can be either positive or negative.
If $t_{\mathrm{ex}}<0$, there is no local minimum value after contact.
However, the parameter condition derived directly from $t_{\mathrm{ex}}<0$ clearly includes the initial states.
In practice, the initial states are unknown when the parameter is designed.
As a result, using the contraposition, a sufficient condition is derived below.
The following equation is given by assuming $t_{\mathrm{ex}}\geq 0$:
\begin{align}
    t_{\mathrm{ex}} &= \frac{1}{\omega_{\mathrm{a}}} + \frac{D_{\mathrm{i}}\nu + K_{\mathrm{i}}\sigma}{(D_{\mathrm{i}}\omega_{\mathrm{a}}- K_{\mathrm{i}})\eta} \nonumber\\
                    &= \frac{(D_{\mathrm{i}}\omega_{\mathrm{a}} - K_{\mathrm{i}})\eta + \omega_{\mathrm{a}}(D_{\mathrm{i}}\nu + K_{\mathrm{i}}\sigma)}{\omega_{\mathrm{a}}(D_{\mathrm{i}}\omega_{\mathrm{a}}-K_{\mathrm{i}})\eta}\geq 0. \label{eq:tex_condition2}
\end{align}
Because $\omega_{\mathrm{a}}>0$, $\eta>0$, and $D_{\mathrm{i}}\omega_{\mathrm{a}} - K_{\mathrm{i}}<0$, \eqref{eq:tex_condition2} means
\begin{equation}
    (D_{\mathrm{i}}\omega_{\mathrm{a}} - K_{\mathrm{i}})\eta + \omega_{\mathrm{a}}(D_{\mathrm{i}}\nu + K_{\mathrm{i}}\sigma) \leq 0.\label{eq:num_tex_condition2}
\end{equation}
Equation \eqref{eq:num_tex_condition2} can be deformed as follows since $\nu>0$, $\omega_{\mathrm{i}}>0$, and $M_{\mathrm{i}}>0$:
\begin{equation}
    4\zeta_{\mathrm{i}}\omega_{\mathrm{a}}-\omega_{\mathrm{i}} \leq -\frac{2\zeta_{\mathrm{i}} \omega_{\mathrm{a}}^2\sigma}{\nu}.\label{eq:deformed_num_tex_condition2}
\end{equation}
Because the right-hand side of \eqref{eq:deformed_num_tex_condition2} is always negative, \eqref{eq:deformed_num_tex_condition2} proves that ``if \eqref{eq:omega_i_condition2} is satisfied, $t_{\mathrm{ex}}\geq 0$ indicates $4\zeta_{\mathrm{i}}\omega_{\mathrm{a}}-\omega_{\mathrm{i}} < 0$, i.e., $\omega_{\mathrm{i}} > 4\zeta_{\mathrm{i}}\omega_{\mathrm{a}}$.''
Therefore, the contraposition, i.e., ``if \eqref{eq:omega_i_condition2} is satisfied, $\omega_{\mathrm{i}} \leq 4\zeta_{\mathrm{i}}\omega_{\mathrm{a}}$ indicates $t_{\mathrm{ex}}<0$'' is true.

Eventually, the parameter condition for the case where $f_{\mathrm{tra}}$ has neither a local maximum nor minimum value after contact is as follows.
\begin{equation}
    2\zeta_{\mathrm{i}}\omega_{\mathrm{a}} < \omega_{\mathrm{i}} \leq 4\zeta_{\mathrm{i}}\omega_{\mathrm{a}}. \label{eq:omega_i_condition_best}
\end{equation}
Note that $\omega_{\mathrm{i}}\leq 4\zeta_{\mathrm{i}}\omega_{\mathrm{a}}$ is conservative.
If the initial states are obtained, the necessary condition can be derived by directly solving $t_{\mathrm{ex}}<0$.
This fact can be confirmed in Fig.~\ref{fig:graph_gen_ftra}.

The following equation given by deforming \eqref{eq:omega_i_condition_best} is often helpful for design.
\begin{equation}
    \frac{\omega_{\mathrm{i}}}{4\zeta_{\mathrm{i}}}\leq \omega_{\mathrm{a}} < \frac{\omega_{\mathrm{i}}}{2\zeta_{\mathrm{i}}}.\label{eq:omega_a_condition_best}
\end{equation}
Note that $\omega_{\mathrm{i}}$ and $\zeta_{\mathrm{i}}$ are based on the estimated $m$ and are influenced by model errors in the impedance control part.
The parameter design with a certain margin is desirable.

%%%%%%%%%%%%%%%%%%%%%%%%%%%%%%%%%%%%%%%%%%%
\section{Experiments for force control-based}
\label{sec:experiments}
%%%%%%%%%%%%%%%%%%%%%%%%%%%%%%%%%%%%%%%%%%%
%=========================================%
\subsection{Experimental Setup}
\label{sec:exp_setup}
%=========================================%
\nonfig{
  \begin{figure}[!t]
    \centering
    \includegraphics[keepaspectratio,width=1.0\linewidth]{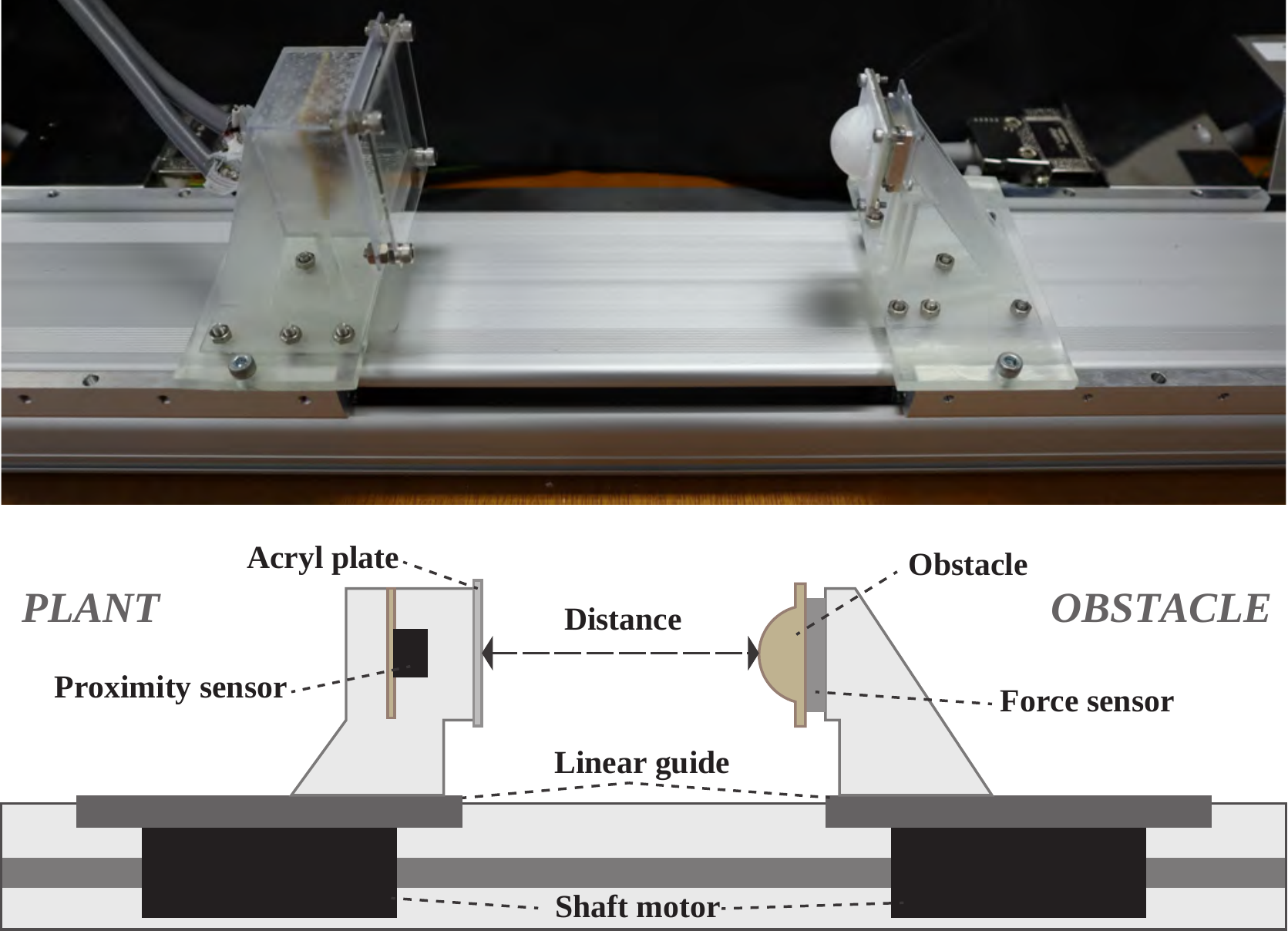}
    \caption{
      The image and diagram of the experimental setup.
      The diagram is in the side view.
      The semitransparent parts are fixtures manufactured with a stereolithography printer.
    }
    \label{fig:experimental_setup}
  \end{figure}
}

\nonfig{
  \begin{figure}[!t]
    \centering
    \includegraphics[keepaspectratio,width=1.0\linewidth]{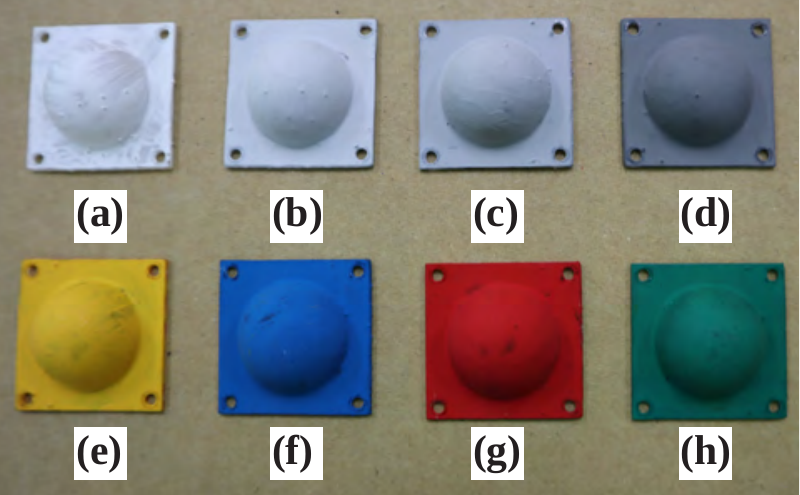}
    \caption{
      The attachments of the obstacle's contact part.
      Their shapes are all the same, half-sphere.
      The description of each color is in Table~\ref{table:obstacle_colors}.
    }
    \label{fig:obstacles}
  \end{figure}
}

\begin{table}
  % increase table row spacing, adjust to taste
  \renewcommand{\arraystretch}{1.3}
  \caption{Obstacle's Colors}
  \label{table:obstacle_colors}
  \centering
  % Some packages, such as MDW tools, offer better commands for making tables
  % than the plain LaTeX2e tabular which is used here.
  \begin{tabular}[t]{ccc}
    \hline
    \begin{tabular}{c}
      Index\\ in Fig.~\ref{fig:obstacles}
    \end{tabular} &
    \begin{tabular}{c}
      Color name\\ in this paper
    \end{tabular} &
    The paint's color name\\
    \hline
    (a) & White & WHITE\\
    (b) & Gray1 & NEUTRAL GRAY8\\
    (c) & Gray2 & NEUTRAL GRAY7\\
    (d) & Gray3 & NEUTRAL GRAY5\\
    (e) & Red & PERMANENT RED\\
    (f) & Blue & SKY BLUE\\
    (g) & Yellow & PERMANENT LEMON\\
    (h) & Green & PEMANENT GREEN MIDDLE\\
    \hline
  \end{tabular}
\end{table}

\noindent
Fig.~\ref{fig:experimental_setup} shows the overview of the experimental setup.
The setup, which includes a proximity sensor, an obstruction, a force sensor for observation, a dual linear stage, and a real-time controller, is capable of realizing the scenarios in Fig.~\ref{fig:situations}.
The dual linear stage~(SLP-15-300-D-M3-A3-SH, NIPPON PULSE MOTOR Co., Ltd.) can individually move two tables.
The plant with the proximity sensor was placed on the left table in Fig.~\ref{fig:experimental_setup}.
The obstacle was created using the right table in Fig.~\ref{fig:experimental_setup}.
Two motor drivers~(MADLT11SM, Panasonic Industry Co., Ltd.) control one shaft motor each.
The motor driver has four modes (position/velocity/force/velocity\&force).
The velocity\&force mode indicates that it can alternate between the velocity and force modes while being used.
The obstacle was manipulated in the experiments using the velocity\&force mode.
These actuators and tables are connected by linear guides.

The force sensor~(USL06-H5-50N-C and DSA-03A, Tec Gihan Co., Ltd.) was used only for measuring the contact force, not to control the plant.
The proximity sensor is developed by parallelly connecting three photo reflectors~(RPR-220, ROHM Co., Ltd.).
The signal-to-noise ratio is improved by the parallel connection.
There is no computational circuit in the rudimentary proximity sensor, which lacks it.
The proximity sensor has an acrylic plate in front of it to prevent obstacles from getting closer than the focal distance or directly touching the sensing element.
By calibrating the proximity sensor, the influence of the acrylic plate on the output was eliminated.

The obstacle's contact part can be replaced with other attachments shown in Fig.~\ref{fig:obstacles}.
These attachments are painted with acrylic paints~(ACRYL GOUACHE, TURNER COLOUR WORKS LTD.) in a variety of colors.
Table~\ref{table:obstacle_colors} lists the names of the paints and colors used in this study.
The obstacle has a half-sphere shape with a radius of $10$~mm, placing the contact point and the proximity sensor's detected point on the force sensor's Z-axis.

The real-time controller~(MicroLabBox, dSPACE GmbH) measures the signal of the proximity sensor, force sensor, and encoder mounted in the dual linear stage and calculates the commands to the motor drivers.
The control frequency is $1$~kHz, whereas the measurement frequency of data for evaluation is $10$~kHz.
The noise on the proximity sensor's output is reduced using a fifth-order Butterworth low-path filter with a cutoff frequency of $500$~Hz.

%=========================================%
\subsection{Evaluation of Impact Reduction}
\label{sec:eva_impact}
%=========================================%
%-----------------------------------%
\subsubsection{Evaluation Method of Impact Reduction}
%-----------------------------------%
\begin{table}
  % increase table row spacing, adjust to taste
  \renewcommand{\arraystretch}{1.3}
  \caption{Control Parameters}
  \label{table:control_parameters}
  \centering
  % Some packages, such as MDW tools, offer better commands for making tables
  % than the plain LaTeX2e tabular which is used here.
  \begin{tabular}[t]{cccccccc}
    \hline
    \multirow{2}{*}{Experiment} & \multicolumn{4}{c}{Impact reduction} & \multicolumn{3}{c}{Contact impedance}\\
    \cline{2-8}
     & $G_{\mathrm{p}}$ & $M_{\mathrm{a}}$ & $\omega_{\mathrm{a}}$ & $\zeta_{\mathrm{a}}$ & $m$ & $\omega_{\mathrm{i}}$ & $\zeta_{\mathrm{i}}$\\
    \hline \hline
    in Section~\ref{sec:eva_impact} & \multirow{2}{*}{0.8} & \multirow{2}{*}{1} & \multirow{2}{*}{5} & \multirow{2}{*}{1} & - & - & - \\
    \cline{1-1}\cline{6-8}
    in Section~\ref{sec:eva_transition} &  &  &  &  & 0.5 & 15 & 1 \\
    \hline
  \end{tabular}  
\end{table}
\noindent
The aim of this experiment is to confirm that the impact reduction method is effective and independent of reflectance.
In this experiment, the impedance control part is replaced with the position controller implemented in the motor driver (the position mode), based on the results of Section~\ref{sec:divided_design} to evaluate without considering the effect of contact impedance control.

The situation in this experiment was the same as shown in Fig.~\ref{fig:situations}(c).
The obstacle approached with a constant velocity of $-0.3$~m/s using the velocity mode of the motor driver.
After making contact, the mode changed to force mode with a force command set to $0$~N, indicating that friction was the only resistance to moving the object.
The plant's desired location $x_{\mathrm{d}}$ was set to $0$~m.
The collision was performed three times for each object shown in Fig.~\ref{fig:obstacles}.
The control parameters in this experiment are shown in Table~\ref{table:control_parameters}.

%-----------------------------------%
\subsubsection{Evaluation Results of Impact Reduction}
%-----------------------------------%
\nonfig{
  \begin{figure}[!t]
    \centering
    \includegraphics[keepaspectratio,width=0.9\linewidth]{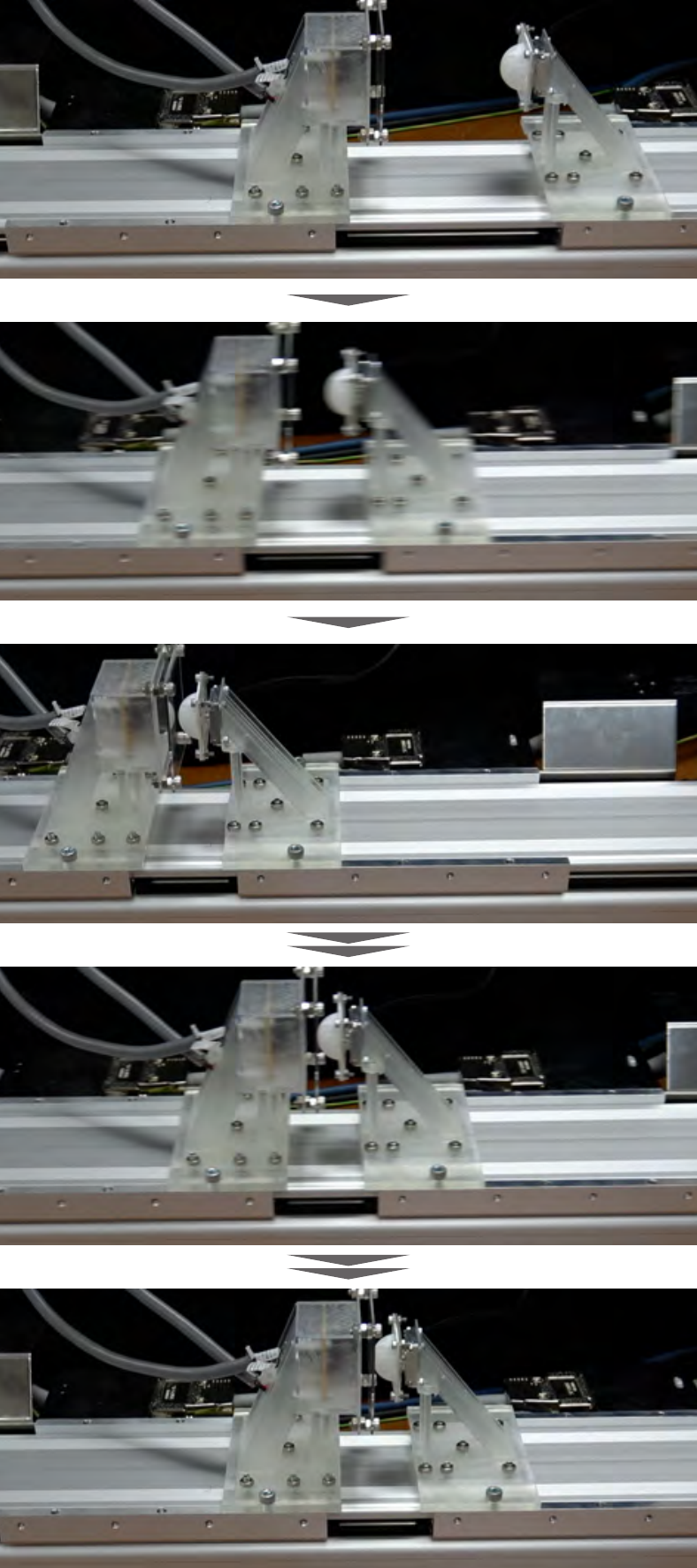}
    \caption{
      The behavior of impact reduction when the obstacle approached the plant.
      The preemptive movement occurred by virtual viscous force.
      After the contact, the plant returned to its origin because the virtual viscous force became zero.
    }
    \label{fig:snap_onlyadm}
  \end{figure}
}

\begin{table*}
  % increase table row spacing, adjust to taste
  \renewcommand{\arraystretch}{1.3}
  \caption{Impact reduction effects when the obstacles of different colors approached}
  \label{table:comparison_color}
  \centering
  % Some packages, such as MDW tools, offer better commands for making tables
  % than the plain LaTeX2e tabular which is used here.
  \begin{tabular}[t]{c r r r r r r r r r }
    \hline
     & \begin{tabular}{c}w/o\\ reduction\\ ($G_{\mathrm{p}}=0.00$)\end{tabular} & White & Gray1 & Gray2 & Gray3 & Yellow & Blue & Red & Green\\
    \hline \hline
    Reflectance ratio to White~(\%) & - & 100 & 76.5 & 54.1 & 30.0 & 81.3 & 78.7 & 76.8 & 63.3 \\
    \hline
    \begin{tabular}{c}Impact reduction\\effect ratio to White~(\%)\end{tabular} & 0.00 & 100 & 98.0 & 94.5 & 81.2 & 96.3 & 97.0 & 96.7 & 97.1 \\
    \begin{tabular}{c}Impact reduction\\ effect~(\%)\end{tabular} & 0.00 & 75.1 & 73.6 & 71.0 & 61.0 & 72.4 & 72.9 & 72.7 & 72.3\\
    \hline
    Ratio of impact force~(\%) & 100 & 24.9 & 26.4 & 29.0 & 39.0 & 27.6 & 27.1 & 27.3 & 27.0 \\ 
    Mean value of impact force~(N) & 15.3 & 3.81 & 4.04 & 4.44 & 5.97 & 4.23 & 4.15 & 4.19 & 4.14 \\
    SD of impact force~(N) & 0.0553 & 0.145 & 0.115 & 0.0671 & 0.266 & 0.0818 & 0.0451 & 0.0538 & 0.115\\
    \hline
    The first impact force~(N) & 15.3 & 3.61 & 4.02 & 4.35 & 5.60 & 4.30 & 4.20 & 4.22 & 3.98 \\
    The second impact force~(N) & 15.3 & 3.88 & 3.90 & 4.45 & 6.07 & 4.28 & 4.09 & 4.23 & 4.21 \\
    The third impact force~(N) & 15.4 & 3.94 & 4.19 & 4.51 & 6.23 & 4.12 & 4.17 & 4.11 & 4.23 \\
    \hline
  \end{tabular}
\end{table*}

\nonfig{
  \begin{figure}[!t]
    \centering
    \includegraphics[keepaspectratio,width=1.0\linewidth]{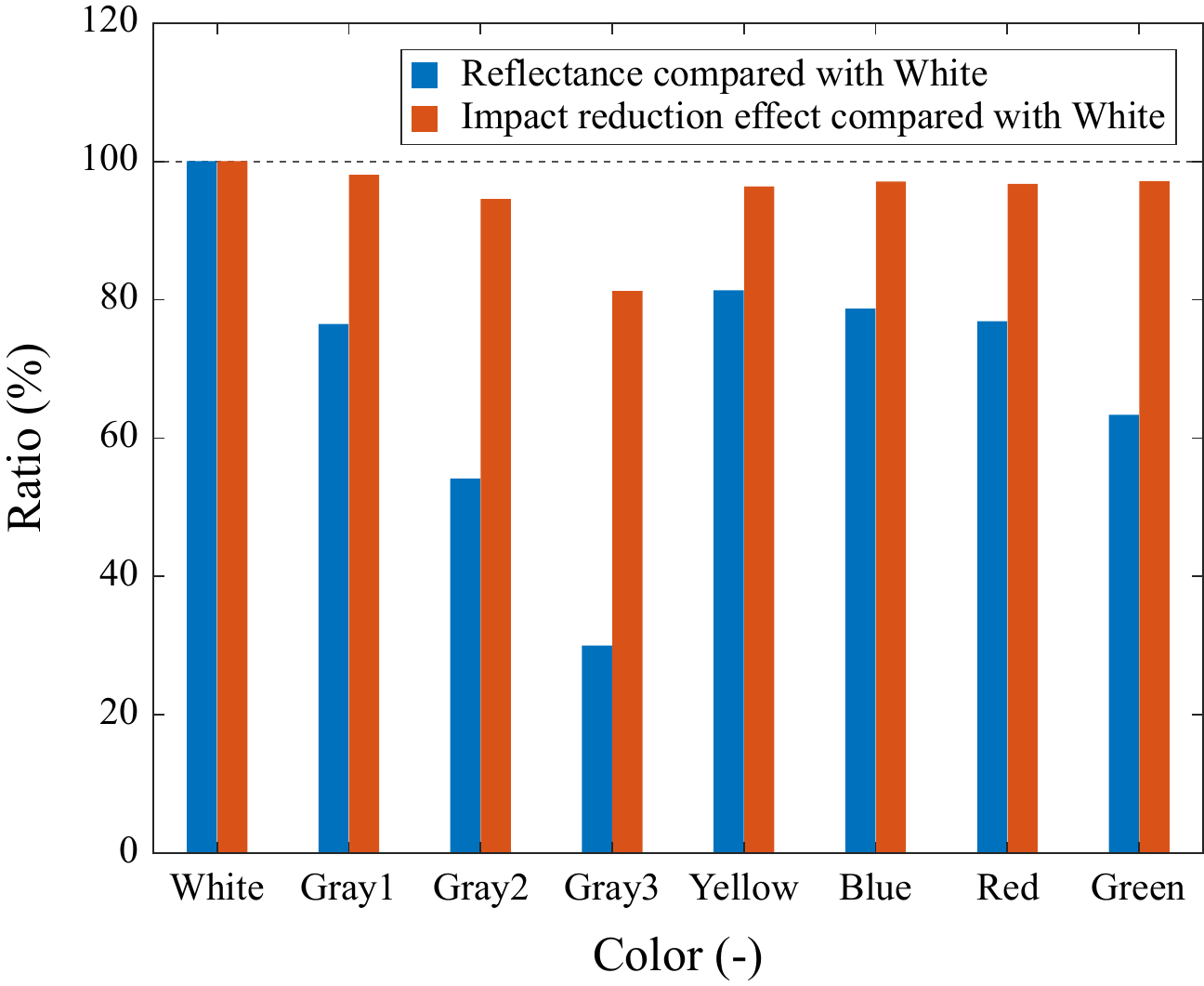}
    \caption{
      The relationship between reflectance and impact reduction effect.
      The differences in impact reduction effect are less than that in reflectance.
    }
    \label{fig:graph_reflectance_impact_reduction}
  \end{figure}
}

\noindent
Fig.~\ref{fig:snap_onlyadm} shows the behavior of impact reduction.
When the impediment was close by, the plant backed away from it.
When the impediment got to the plant, the plant forced the obstruction to return to its starting point because $f_{\mathrm{p}}$ turned to $0$.
The behavior during going back to the origin is the free vibration of the virtual object decided in the admittance control part.

The impact force measurements are shown in Table~\ref{table:comparison_color}.
Note that the maximum force sensor output is the impact force.
It is confirmed that the peak outputs happened right after contact.

When $G_{\mathrm{p}}=0$, i.e., the impact reduction was disabled, the mean value of impact force was $15.3$~N.
The mean values of impact force were $3.81 \text{--} 5.97$~N when the impact reduction was enabled.
The impact reduction effect was $61.0 \text{--} 75.1$~\%, where the impact reduction effect was calculated as follows: $100 \times$[(the mean without reduction) - (the mean of the case)]/(the mean without reduction).

Moreover, the impact reduction effect ratios were computed by normalizing with the impact reduction effect for White to confirm the relationship between reflectance and the reduction effect.
The reflectance ratio was calculated by dividing each proximity sensor's output during contact by the output for White.
Fig.~\ref{fig:graph_reflectance_impact_reduction} shows the link.

Fig.~\ref{fig:graph_reflectance_impact_reduction} shows that reflectance has no influence on the impact reduction effect.
The difference in the decrease impact was only $5.5$~\% even though Gray2 had a reflectance that was half that of White.
Gray3 reduces the reduction effect, nevertheless, to $81.3$~\%.
The drop results from the calibration being insufficient.
Although the proximity sensor cannot detect any obstacles, there is a very faint signal in the output after calibration.
The residual signal is dominant when the proximity sensor output is extremely small and suppresses the virtual viscous force to comply with \eqref{eq:fp}.
The leftover signal ultimately prevents the behavior from decreasing impact.
Unlike the reflectance problem, this problem can be solved by increasing the amount of light emitted or the sensitivity of the sensor.

%=========================================%
\subsection{Evaluation of Smooth Transition}
\label{sec:eva_transition}
%=========================================%
%-----------------------------------%
\subsubsection{Evaluation Method of Smooth Transition}
%-----------------------------------%
\noindent
This experiment was conducted to ensure that impact reduction is not limited by contact impedance control and that the transition from impact reduction to contact impedance control is seamless.

First, the collision shown in Fig.~\ref{fig:situations}(c) was executed.
Obstacle velocity was set to $-0.3$~m/s for comparison with Section~\ref{sec:eva_impact} and $-0.4$~m/s to examine the influence of different obstacle velocities.

The experiment was then performed under the scenario depicted in Fig.~\ref{fig:situations}(d).
The plant followed the desired state and ran into the roadblock placed in its path.
The trajectory planner generated the desired state trajectory to be a minimum jerk trajectory.

In these experiments, the only color of the obstacle was White.
The impedance control part was implemented using \eqref{eq:input_imp3}, and the control settings were established as stated in Table~\ref{table:control_parameters}.
Note that $m$ was decided from the datasheet of the dual linear stage, and $\zeta_{\mathrm{i}} = 1$ and $\omega_{\mathrm{i}} = 3\omega_{\mathrm{a}}$ such that $\omega_{\mathrm{a}}$, $\zeta_{\mathrm{i}}$, and $\omega_{\mathrm{i}}$ satisfies \eqref{eq:omega_i_condition_best}.

%-----------------------------------%
\subsubsection{Evaluation Results of Smooth Transition}
%-----------------------------------%
\begin{table*}
  % increase table row spacing, adjust to taste
  \renewcommand{\arraystretch}{1.3}
  \caption{Impact reduction effects of contact impedance alone and combined noncontact \& contact impedance control}
  \label{table:imp_vs_admimp}
  \centering
  % Some packages, such as MDW tools, offer better commands for making tables
  % than the plain LaTeX2e tabular which is used here.
  \begin{tabular}[t]{c p{60pt} p{60pt} p{60pt} p{60pt}}
   \hline
   \begin{tabular}{c}Enabled impedance control\end{tabular} & \multicolumn{2}{r}{\begin{tabular}{c} Contact ($G_{\mathrm{p}}=0.00$) \end{tabular}} & \multicolumn{2}{r}{\begin{tabular}{c}Noncontact \& Contact ($G_{\mathrm{p}}=0.80$)\end{tabular}}\\
   \hline
   \begin{tabular}{c}Obstacle velocity~(m/s)\end{tabular} &\hfill -0.3 \hfill &\hfill -0.4  &\hfill -0.3  &\hfill -0.4 \\
   \hline\hline
   \begin{tabular}{c}Impact reduction effect ratio to\\ w/o reduction shown in Table~\ref{table:comparison_color}~(\%)\end{tabular} & \hfill 17.9 &\hfill -  & \hfill 76.8  &\hfill - \\
   \begin{tabular}{c}Ratio of impact force to\\ w/o reduction shown in Table~\ref{table:comparison_color}~(\%)\end{tabular} & \hfill 82.1 & \hfill - & \hfill 23.2 & \hfill - \\
   \hline
   \begin{tabular}{c}Impact reduction effect ratio to $G_{\mathrm{p}}=0$~(\%)\end{tabular} & \hfill 0.00 &\hfill 0.00  & \hfill 71.7  &\hfill 74.1 \\
   \hline
   \begin{tabular}{c}Ratio of impact force to $G_{\mathrm{p}}=0$~(\%)\end{tabular} & \hfill 100 & \hfill 100 & \hfill 23.2 & \hfill 25.9 \\
   Mean value of impact force~(N) & \hfill 12.6 & \hfill 17.5 & \hfill 3.56 & \hfill 4.53\\
   SD of impact force~(N) & \hfill 0.0687 & \hfill 0.0410 & \hfill 0.0988 & \hfill0.155\\
   \hline
   The first impact force~(N) & \hfill 12.5 & \hfill 17.5 & \hfill 3.65 & \hfill 4.40\\
   The second impact force~(N) & \hfill 12.6 & \hfill 17.5 & \hfill 3.61 & \hfill 4.75\\
   The third impact force~(N) & \hfill 12.6 & \hfill 17.6 & \hfill 3.42 & \hfill 4.45\\
   \hline
 \end{tabular}
\end{table*}

\nonfig{
  \begin{figure}[!t]
    \centering
    \includegraphics[keepaspectratio,width=1.0\linewidth]{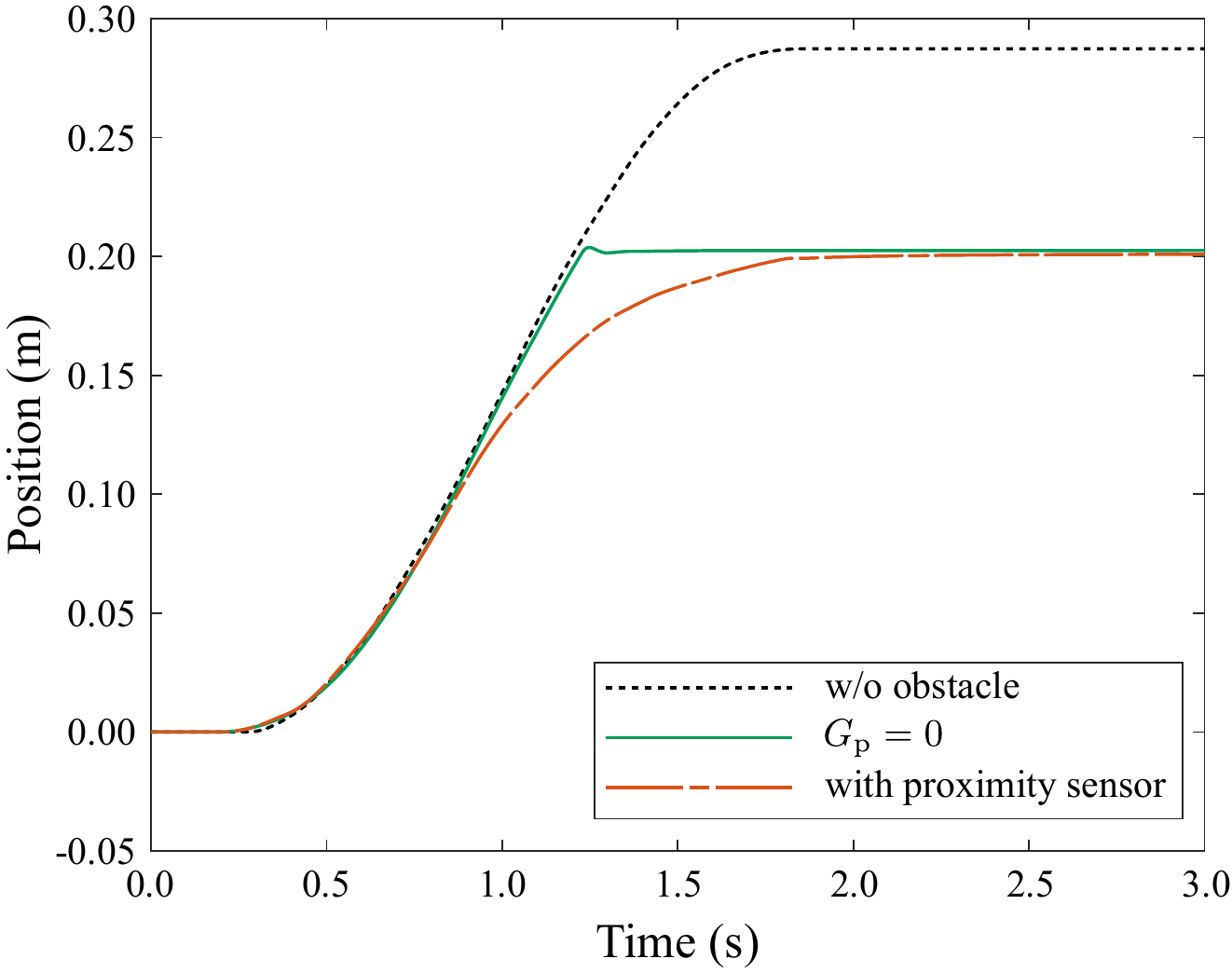}
    \caption{
      Position trajectories of the plant when it followed intended course.
      The differences in these trajectories were brought about by the fixed obstacle and preemptive impact reduction feature.
    }
    \label{fig:graph_position_trajectory}
  \end{figure}
}

\nonfig{
  \begin{figure}[!t]
    \centering
    \includegraphics[keepaspectratio,width=1.0\linewidth]{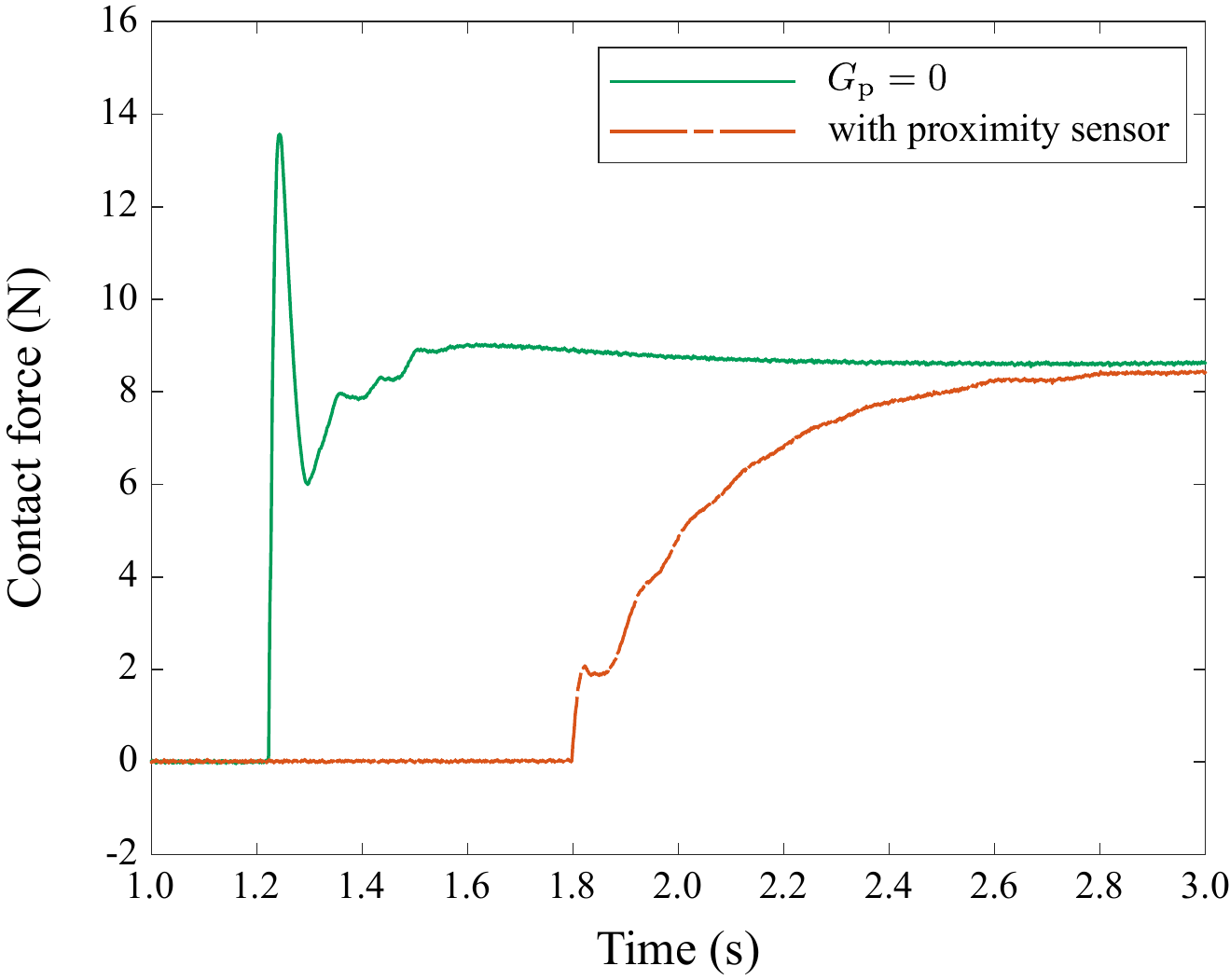}
    \caption{
      The measured contact force that occurred when the plant hit the fixed obstacle on the path.
      Each impact force is shown by the first peak of each wave.
      Before convergence in the outcome, when the preemptive impact reduction was enabled, there was no peak other than the impact force.
    }
    \label{fig:graph_force_trajectory}
  \end{figure}
}

\noindent
Table~\ref{table:imp_vs_admimp} shows the results where the obstacle approached the plant.
When the obstacle velocity was $-0.3$~m/s, the mean value of impact force was $12.6$~N when $G_{\mathrm{p}}=0$.
By controlling contact impedance, the impact force is decreased.
As per Section \ref{sec:eva_impact}, the impact reduction effect ratio to no reduction was $17.9$~\%.
The proximity sensor was turned on, and the average impact force was $3.56$~N.
The impact reduction effect ratio to without reduction shown in Section~\ref{sec:eva_impact} was $76.8$~\%, and the one to $G_{\mathrm{p}}=0$ was $71.7$~\%.
These results indicate that the combination of preemptive impact reduction and contact impedance control is adequate for impact reduction.
Preemptive impact reduction alone, as indicated in Section~\ref{sec:eva_impact}, and contact impedance control alone both had lower effects than the combined controller.

Under the condition of obstacle velocity was $-0.4$~m/s, the mean value of impact force was $17.5$~N for $G_{\mathrm{p}}=0$, whereas it was $4.53$~N with the proximity sensor.
The impact reduction effect ratio to $G_{\mathrm{p}}=0$ was $74.1$~\%.
This result shows that the effect of impact reduction was equal to or greater than that for $-0.3$~m/s.

Fig.~\ref{fig:graph_position_trajectory} shows the plant's positional trajectory when it followed the desired state.
When there was no barrier, the plant continued along the minimum jerk trajectory that was intended.
When the trajectory encountered a fixed impediment, the plant collided and came to a stop.
Before the accident, the plant was slowed by the proximity sensor's impact reduction.
Note that since the results of the three trials were almost the same, the figure depicts only one of each.

The measured contact forces are shown in Fig.~\ref{fig:graph_force_trajectory}.
Note that the horizontal axis scale is different from that in Fig.~\ref{fig:graph_position_trajectory}.
The peak of the impact force was $13.6$~N for $G_{\mathrm{p}}=0$, whereas it was $2.08$~N with the proximity sensor.
The impact reduction effect ratio to $G_{\mathrm{p}}=0$ was $84.7$~\%.
Moreover, when the proximity sensor was activated, as shown analytically in Section~\ref{sec:smooth_transition}, the contact force gradually increased.
No undesired contact force was present right away following the incident.
The length of the transition is determined by $\omega_{\mathrm{a}}$.
This outcome demonstrated that the transition from impact reduction to contact impedance control is smooth.

%%%%%%%%%%%%%%%%%%%%%%%%%%%%%%%%%%%%%%%%%%%
\section{Expansions}
\label{sec:expansions}
%%%%%%%%%%%%%%%%%%%%%%%%%%%%%%%%%%%%%%%%%%%
%=========================================%
\subsection{Proposed Controller for Position/Velocity Control-based Robots}
\label{sec:proposal_posvel}
%=========================================%
\nonfig{
  \begin{figure*}[!t]
    \centering
    \includegraphics[keepaspectratio,width=1.0\linewidth]{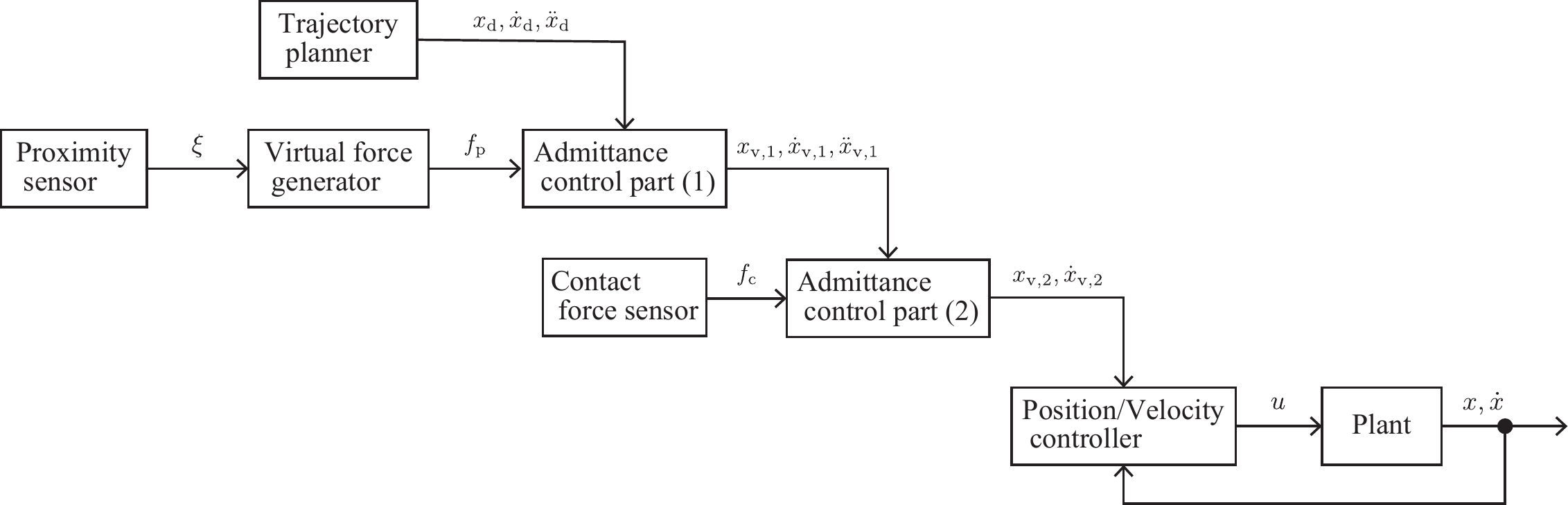}
    \caption{
      Block diagram of the proposed controller for position/velocity control-based robots.
      There are two admittance control parts.
      The other is for contact force, whereas the first is for virtual force.
      The output of the first admittance control part serves as the second admittance control part's desired state.
    }
    \label{fig:block_diagram_posvel}
  \end{figure*}
}

\noindent
Fig.~\ref{fig:block_diagram_posvel} shows the proposed controller for position/velocity control-based robots.
The second admittance control part and position/velocity controller are used in place of the impedance control part in contrast to Fig.~\ref{fig:block_diagram_force}.
In the following, the controller shown in Fig.~\ref{fig:block_diagram_force} is referred to as PACIC (Proximity Admittance and Contact Impedance Controller), whereas the controller shown in Fig.~\ref{fig:block_diagram_posvel} is referred to as PACAC (Proximity Admittance and Contact Admittance Controller).

In PACAC, \eqref{eq:dynamics_adm} is replaced with
\begin{equation}
    M_{\mathrm{a},1} (\ddot{x}_{\mathrm{v},1} - \ddot{x}_{\mathrm{d}}) + D_{\mathrm{a},1}(\dot{x}_{\mathrm{v},1} - \dot{x}_{\mathrm{d}}) + K_{\mathrm{a},1}(x_{\mathrm{v},1} - x_{\mathrm{d}}) = f_{\mathrm{p}}, \label{eq:dynamics_adm_1}
\end{equation}
where the subscripts are expanded from each.
The following equation and any position/velocity controller are used instead of \eqref{eq:input_imp}:
\begin{align}
    M_{\mathrm{a},2} (\ddot{x}_{\mathrm{v},2} - \ddot{x}_{\mathrm{v},1}) + D_{\mathrm{a},2}(\dot{x}_{\mathrm{v},2} - \dot{x}_{\mathrm{v},1})&\nonumber\\
     + K_{\mathrm{a},2}(x_{\mathrm{v},2} - x_{\mathrm{v},1}) &= f_{\mathrm{c}}. \label{eq:dynamics_adm_2}
\end{align}
By replacing \eqref{eq:dynamics_adm} with \eqref{eq:dynamics_adm_1} and \eqref{eq:dynamics_imp} with \eqref{eq:dynamics_adm_2}, the same analysis results as in Section~\ref{sec:divided_design} and Section~\ref{sec:smooth_transition} are obtained.
The characteristic difference between PACIC and PACAC is similar to the distinction between conventional admittance and impedance control mentioned in Section~\ref{sec:rel_contact_impedance}.

%=========================================%
\subsection{Generalized Serial Combined Impedance Controller}
\label{sec:proposal_generalized}
%=========================================%
\nonfig{
  \begin{figure*}[!t]
    \centering
    \includegraphics[keepaspectratio,width=1.0\linewidth]{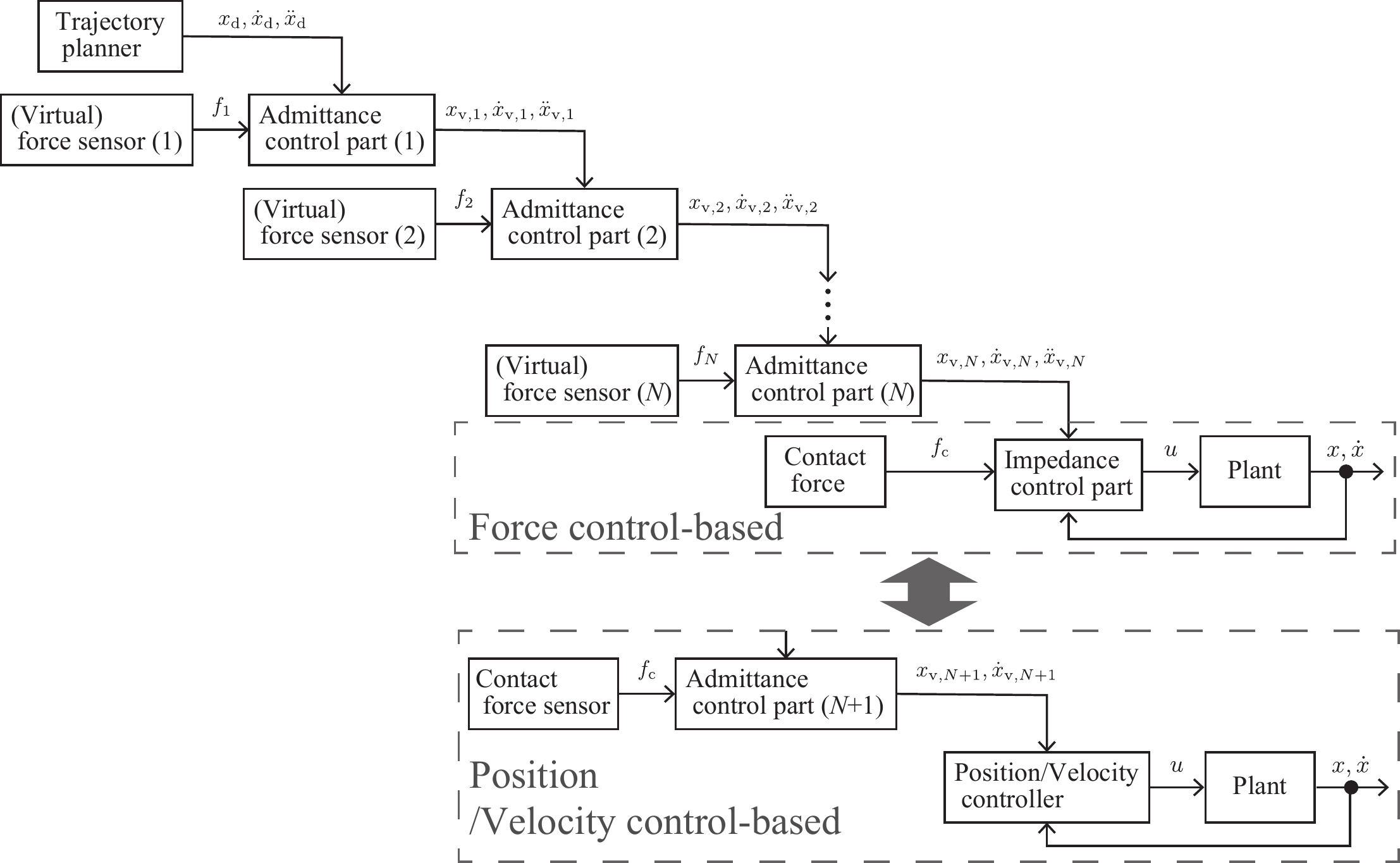}
    \caption{
      The generalized serial combined impedance controller is shown in a block diagram.
      As with the controller in \cite{fujiki2021numerical}, the same virtual or contact force can also serve as the input for several admittance control parts.
      Depending on the plant's control type, the last part should be chosen.
    }
    \label{fig:block_diagram_generalized}
  \end{figure*}
}

\noindent
Fig.~\ref{fig:block_diagram_generalized} shows the generalized controller, including the controller proposed in \cite{fujiki2021numerical}, PACIC, and PACAC.
There are $N$ admittance control parts, as well as either an additional admittance control part or an impedance control part.
Virtual or contact force sensor outputs are received by each admittance control part.
The following part's desired state is based on each output of the admittance control part.
Each admittance control part determines the influence of the related sensor output on its target state, as discussed in Section~\ref{sec:divided_design}.
Keep in mind that delays could happen depending on the control parameters.
The stacked delay will undermine the performance of the control.

The controller proposed in \cite{fujiki2021numerical} has an admittance control part and an impedance control part, and the admittance control part receives contact force sensor output.
Namely, in PACIC and PACAC, two control parts are connected through the virtual object's state, whereas in the controller presented in \cite{fujiki2021numerical}, they are also connected through force.
The distinction allows for drastically different controller characteristics from the one given in this study, i.e., the generalized controller may be widely applicable.
The characteristics in the case of \cite{fujiki2021numerical} are further investigated in another paper.

%%%%%%%%%%%%%%%%%%%%%%%%%%%%%%%%%%%%%%%%%%%
\section{Discussion}
\label{sec:discussion}
%%%%%%%%%%%%%%%%%%%%%%%%%%%%%%%%%%%%%%%%%%%
%=========================================%
\subsection{Example of the Design Process}
\label{sec:dis_design_process}
%=========================================%
\noindent
An example of the design process of the proposed controllers is shown below.
\begin{enumerate}
    \item \textit{Parameter design for contact impedance control}: \mbox{}\\
                Design the parameters for either the impedance control part or the second admittance control part as proposed in Section~\ref{sec:proposal_force} or Section~\ref{sec:proposal_posvel} following a given task.
                Nothing needs to be changed if contact impedance/admittance control has been put in place.
    \item \textit{Parameter design for smooth transition}: \mbox{}\\
                Design the parameters of the admittance control part receiving virtual viscous force, so that \eqref{eq:omega_a_condition_best} is satisfied and $\zeta_{\mathrm{a}}=1$.
                The parameter design with some margin is recommended.
    \item \textit{Gain tuning for impact reduction}: \mbox{}\\
                Find the suitable value of $G_{\mathrm{p}}$ in \eqref{eq:fp} for impact reduction.
                This process is ad hoc yet.
                The systematic design method needs to be developed in future work.
\end{enumerate}

%=========================================%
\subsection{Limitations}
\label{sec:dis_limitations}
%=========================================%
\noindent
An inherent limitation of the proposed method is that it requires a proximity sensor.
As stated in \cite{guadarrama2022preemptive}, it is difficult to incorporate a proximity sensor in the contact area.
However, a primitive optical proximity sensor is enough for the proposed method.
Due to their little size, simple optical proximity sensors can be mounted in tight spaces or used to cover contact areas.
There are some examples, such as mounting on a fingertip~\cite{johnston1973optical}, on tiptoe~\cite{sato2022pre}, on feet~\cite{guadarrama2022preemptive}, and covering a robot arm~\cite{cheung1989development}.
Additionally, high-density implementation is also feasible, as presented in~\cite{suzuki2021proximity}, and enhances the signal-to-noise ratio described in Section~\ref{sec:eva_impact}.
Intelligent proximity sensors that measure distance, such as a ToF sensor IC~(VL6180X, STMicroelectronics N.V.), which have grown popular in recent years, and a bespoke sensor~\cite{koyama2019high} can also be utilized in the suggested technique instead of basic proximity sensors.
The virtual viscous force can be derived by differentiating the distance output as proposed in~\cite{koyama2019high}.
Because the admittance control part functions as a reliable second-order filter, it should be noted that even though the differentiator amplifies high-frequency noise generally, the influence of the noise on plant motion is minimal in the proposed method, as demonstrated by the experimental results.
Other forces that satisfy the analysis's presumption that $f_{\mathrm{p}}=0$ in contact can be used in place of the virtual viscous force.
For instance, the virtual elastic force is suitable to the suggested method when using the sensor design method given in~\cite{sato2022pre}.

The plant in this paper is only a 1D mass model.
Future research should think about adapting approaches to complex systems, including robotic hands and legs.
To talk about the collision, nevertheless, investigation and evaluation of the component of the approaching direction are crucial.
The findings of this study will be useful for subsequent work.

%=========================================%
\subsection{(By-product) Contact Preservation}
\label{sec:dis_contact_preservation}
%=========================================%
\noindent
The contact state can also be maintained by the virtual viscous force when the transition from contact to noncontact is likely to happen.
For specific jobs, like keeping a grabbed object from being dropped from a robot hand, this functionality could be useful.
Future study will evaluate contact preservation because doing so requires defining a job.
Keep in mind that if it is not necessary, you can disable the contact preservation feature by using saturation.

%%%%%%%%%%%%%%%%%%%%%%%%%%%%%%%%%%%%%%%%%%%
\section{Conclusion}
\label{sec:conclusion}
%%%%%%%%%%%%%%%%%%%%%%%%%%%%%%%%%%%%%%%%%%%
\noindent
This paper introduces novel control methods that reduce impact and control contact force after a collision.
One of the techniques is for force control-based robots and uses a virtual viscous force generator, an admittance control part, and an impedance control part.
Another employs the second admittance control part for contact force instead of the previously described impedance control part and is for position/velocity control-based robots.

The analyses and experiments verified three advantages.
The first is that even though the impact reduction technique uses a primitive optical proximity sensor to react before contact, the effect of reflectance on impact reduction is tiny.
Despite a little effect brought on by the experiment's insufficient calibration, the analytical outcome showed independence.
The second is that impact reduction and contact impedance control can be accomplished separately by designing the control parameters of the suggested approach.
Preemptive impact reduction is simply added; if contact impedance control is already in place, its parameters are unaltered.
In the experiments, adding the preemptive impact reduction reduced impact force by $71.7$~\% when the obstacle approached the plant and by $84.7$~\% when the plant approached the fixed obstacle.
The third is the smooth transition from impact reduction to contact impedance control.
Both switching and unwelcome touch force are absent.
The contact impedance control is gradually enabled.
The majority of robots that need to interact with environments, objects, and humans can benefit from these advantages.

\appendices
%%%%%%%%%%%%%%%%%%%%%%%%%%%%%%%%%%%

% % you can choose not to have a title for an appendix
% % if you want by leaving the argument blank
% %%%%%%%%%%%%%%%%%%%%%%%%%%%%%%%%%%%
% \section{}
% %%%%%%%%%%%%%%%%%%%%%%%%%%%%%%%%%%%
% Appendix two text goes here.

% use section* for acknowledgment
%%%%%%%%%%%%%%%%%%%%%%%%%%%%%%%%%%%
\section*{Acknowledgment}
%%%%%%%%%%%%%%%%%%%%%%%%%%%%%%%%%%%
This work was partially supported by JSPS KAKENHI Grant Number JP20K14702.
The authors would like to thank Enago~(www.enago.jp) for the English language review.

% Can use something like this to put references on a page
% by themselves when using endfloat and the captionsoff option.
\ifCLASSOPTIONcaptionsoff
  \newpage
\fi

% trigger a \newpage just before the given reference
% number - used to balance the columns on the last page
% adjust value as needed - may need to be readjusted if
% the document is modified later
%\IEEEtriggeratref{8}
% The "triggered" command can be changed if desired:
%\IEEEtriggercmd{\enlargethispage{-5in}}

% references section

% can use a bibliography generated by BibTeX as a .bbl file
% BibTeX documentation can be easily obtained at:
% http://mirror.ctan.org/biblio/bibtex/contrib/doc/
% The IEEEtran BibTeX style support page is at:
% http://www.michaelshell.org/tex/ieeetran/bibtex/
\bibliographystyle{IEEEtran}
% argument is your BibTeX string definitions and bibliography database(s)
\bibliography{IEEEabrv,bib_paper_PACI2022}
%
% <OR> manually copy in the resultant .bbl file
% set second argument of \begin to the number of references
% (used to reserve space for the reference number labels box)
% \begin{thebibliography}{1}

% \bibitem{IEEEhowto:kopka}
% H.~Kopka and P.~W. Daly, \emph{A Guide to \LaTeX}, 3rd~ed.\hskip 1em plus
%   0.5em minus 0.4em\relax Harlow, England: Addison-Wesley, 1999.

% \end{thebibliography}

% biography section
% 
% If you have an EPS/PDF photo (graphicx package needed) extra braces are
% needed around the contents of the optional argument to biography to prevent
% the LaTeX parser from getting confused when it sees the complicated
% \includegraphics command within an optional argument. (You could create
% your own custom macro containing the \includegraphics command to make things
% simpler here.)
%\begin{IEEEbiography}[{\includegraphics[width=1in,height=1.25in,clip,keepaspectratio]{mshell}}]{Michael Shell}
% or if you just want to reserve a space for a photo:

\begin{IEEEbiography}{Hikaru Arita}
  Biography text here.
  % He received his B.E., M.E., and Ph.D. degrees from the University of Electro-Communications in 2012, 2014, and 2019.
  % From 2014 to 2016, he worked at OMRON corporation.
  % From 2019 to 2022, he was an assistant professor at Ritsumeikan University.
  % Since 2022, he has been an assistant professor at Kyushu University.
  % His research interests include proximity sensors and sensor-based control.
\end{IEEEbiography}

\begin{IEEEbiography}{Hayato Nakamura}
  Biography text here.
\end{IEEEbiography}

\begin{IEEEbiography}{Takuto Fujiki}
  Biography text here.
\end{IEEEbiography}

\begin{IEEEbiography}{Kenji Tahara}
  Biography text here.
\end{IEEEbiography}

% % if you will not have a photo at all:
% \begin{IEEEbiographynophoto}{John Doe}
% Biography text here.
% \end{IEEEbiographynophoto}

% % insert where needed to balance the two columns on the last page with
% % biographies
% %\newpage

% \begin{IEEEbiographynophoto}{Jane Doe}
% Biography text here.
% \end{IEEEbiographynophoto}

% You can push biographies down or up by placing
% a \vfill before or after them. The appropriate
% use of \vfill depends on what kind of text is
% on the last page and whether or not the columns
% are being equalized.

%\vfill

% Can be used to pull up biographies so that the bottom of the last one
% is flush with the other column.
%\enlargethispage{-5in}

% that's all folks
\end{document}